\documentclass[journal]{IEEEtran}
\usepackage{stfloats}
\usepackage{amsmath}
\usepackage{amssymb}
\usepackage{textcomp,booktabs}
\usepackage[dvipsnames, svgnames, x11names]{xcolor}
\usepackage{wrapfig}
\usepackage{colortbl,booktabs}
\usepackage{verbatim}
\usepackage{epstopdf}
\usepackage[colorlinks,linkcolor=blue]{hyperref}

\usepackage{floatrow}

\usepackage{booktabs,multirow,rotating}
\usepackage{cite}

\usepackage{subfigure}
\usepackage{bm}
\usepackage{caption}

\usepackage{float}

\makeatletter
\renewcommand{\@thesubfigure}{\hskip\subfiglabelskip}
\makeatother

\usepackage{colortbl}
\definecolor{mygray}{gray}{.9}
\usepackage{diagbox}

\begin{document}
%

\title{A Deeper Look at Salient Object Detection: \\Bi-stream Network with a Small Training Dataset}


\author{Zhenyu Wu$^{1}$  ~~~~~~Shuai Li$^{1}$
 ~~~~~~Chenglizhao Chen$^{1,2*}$\thanks{Corresponding author: Chenglizhao Chen, cclz123@163.com.}  ~~~~~~Hong Qin$^3$ ~~~~~~Aimin Hao$^{1}$\\

$^1$State Key Laboratory of Virtual Reality Technology and Systems, Beihang University\\
$^2$Qingdao University~~~~~~$^3$Stony Brook University\\

%
}

\markboth{IEEE Transactions on Multimedia, VOL.XX, NO.XX, XXX.XXXX}%
{Shell \MakeLowercase{\textit{et al.}}: Bare Demo of IEEEtran.cls for Journals}

\maketitle

\IEEEtitleabstractindextext{
\begin{abstract}
Compared with the conventional hand-crafted approaches, the deep learning based methods have achieved tremendous performance improvements by training exquisitely crafted fancy networks over large-scale training sets. However, do we really need large-scale training set for salient object detection (SOD)? In this paper, we provide a deeper insight into the interrelationship between the SOD performances and the training sets.
To alleviate the conventional demands for large-scale training data, we provide a feasible way to construct a novel small-scale training set, which only contains 4K images.
Moreover, we propose a novel bi-stream network to take full advantage of our proposed small training set, which is consisted of two feature backbones with different structures, achieving complementary semantical saliency fusion via the proposed gate control unit.
To our best knowledge, this is the first attempt to use a small-scale training set to outperform state-of-the-art models which are trained on large-scale training sets;
nevertheless, our method can still achieve the leading state-of-the-art performance on five benchmark datasets. Both the code and dataset are publicly available at \emph{\url{https://github.com/wuzhenyubuaa/TSNet}}.
\end{abstract}


}
\maketitle
\IEEEdisplaynontitleabstractindextext
\IEEEpeerreviewmaketitle


\section{Introduction}

\IEEEPARstart{S}alient object detection (SOD) aims to estimate the most attractive regions of images or videos.
As the pre-processing tool, SOD plays an important role in a wide range of computer vision, such as visual tracking \cite{OurPR15,ChenPR16}, object retargeting~\cite{vinyals2015show}, RGB-D completion~\cite{CC2020TIP}, image retrieval \cite{7004809} and visual question answering~\cite{lin2017task}.

Inspired by cognitive psychology and neuroscience, the classical SOD models~\cite{itti1998model,8554115,6331539,OurTIP15} are developed by fusing various saliency cues, however, all these cues fail to capture the wide variety of visual features regarding the salient objects.
After entering the deep learning era, the SOD performance has achieved tremendous improvement because of both the exquisitely crafted fancy network architectures~\cite{MDF,DSS,RADF} and the availability of large-scale well-annotated training data~\cite{cheng2015global,wang2017learning}.

\begin{figure}[t]
\centering
\includegraphics[width=1\linewidth]{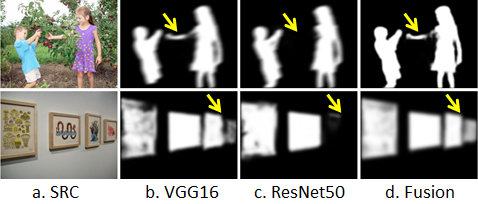}
\caption{Deep features in networks with different architectures are generally complementary, in which these feature maps are obtained from the last convolutional layers.}
\vspace{-0.2em}
\label{fig1}
\end{figure}

Following the single-stream network structure, the most recent SOD methods~\cite{CC2019TMM1,DSS,RADF} have focused on how to effectively aggregate multi-level visual feature maps to boost their performances.
Though remarkable progress has been achieved, these methods have reached their performance bottleneck, because their single-stream structures usually consist of single feature backbone, which usually results in limited semantical sensing ability.
Theoretically, different network architectures have inequable feature response even if for same image.
As a result, we may easily achieve complementary semantical deep features if we simultaneously use two distinct feature backbones, please refer to the pictorial demonstrations in Fig.~\ref{fig1}.

\begin{figure*}[t]
\centering
\includegraphics[width=1\linewidth]{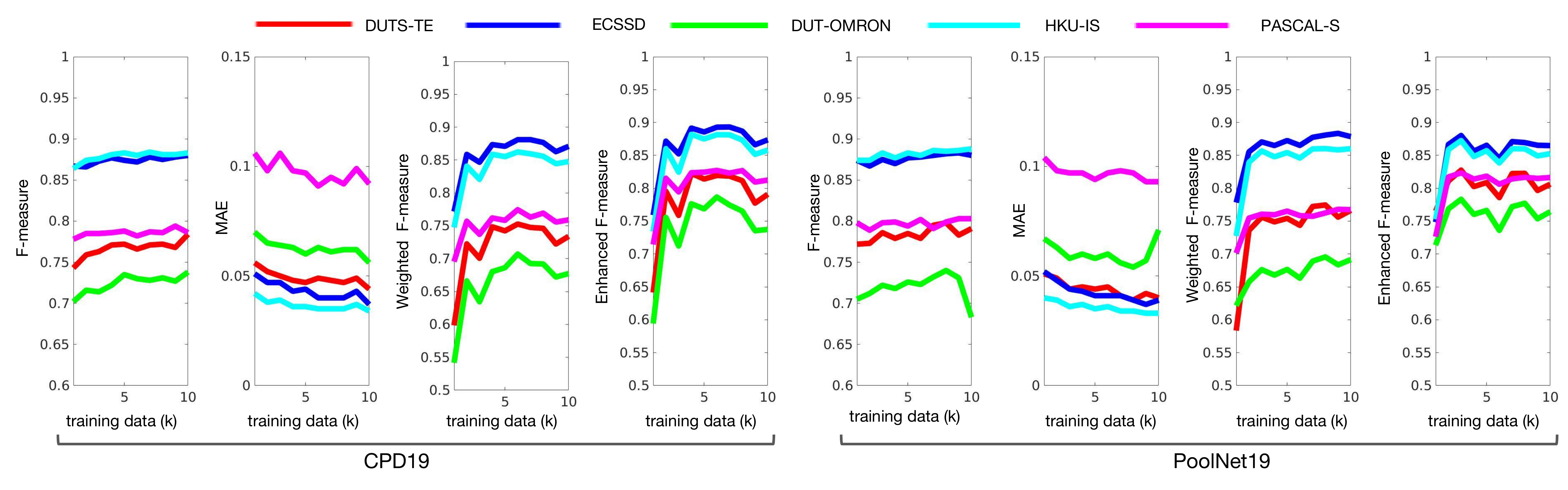}
\caption{The quantitative performances of 2 state-of-the-art models (CPD19~\cite{CPD} and PoolNet19~\cite{PoolNet}) vary with the training data size, showing that the conventional consensus regarding the relationship between the model performance and the training set size---``\emph{the model performance is positively related to the training set size}'' may not always hold.}
\vspace{-0.5cm}
\label{data_analysis_fig2}
\end{figure*}

In terms of the training dataset, the SOD community has reached a consensus on the training protocol, i.e., trained on the MSRA10K~\cite{cheng2015global} or DUTS-TR~\cite{wang2017learning} dataset, and then tested on other datasets.
However, is this training strategy the best choice?
According to our experimental results, some inspiring findings can be summarized as follows:
1) The overall model performance is not always positively correlated with the number of training data, see the quantitative proofs in Fig. {\ref{data_analysis_fig2}};
2) The performances of deep models trained on single training dataset (MSRA10K or DUTS-TR) are usually limited due to the unbalanced semantic distribution problem, as evidenced in Fig. {\ref{data_analysis_fig1}}; (3) The MSRA10K and the DUTS-TR datasets are complementary.

From the perspective of neuroscience, the human visual system comprises two largely independent subsystems that mediate different classes of visual behaviors~\cite{visualParallel,schiller1991parallel}.
The subcortical projection from the retina to cerebral cortex is strongly dominated by the two pathways that are relayed by the magnocellular (M) and parvocellular (P) subdivisions of the lateral geniculate nucleus (LGN).
Parallel pathways generally exhibit two main characteristics:
1) The M cells contribute to transient processing (e.g., visual motion perception, eye movement, etc.) while the P cells contribute more to recognition (e.g., object recognition, face recognition, etc.);
2) The M and P cells are separated in the LGN, but it is recombined in visual cortex latter.

Inspired by the above-mentions, we first build a semantic category balanced small-scale training dataset namely MD4K (total 4172 images) from the off-the-shelf MSRA10K and DUTS-TR datasets.
To take full advantage of the proposed small training set, we then propose a novel bi-stream network, consisting of two sub-branches with different network structures, which aims to explore complementary semantical information to obtain more powerful feature representation for the SOD task.
Meanwhile, we devise a novel gate control unit to effectively fuse complementary information encoded in different sub-branches.
Moreover, we introduce the multi-layer attention into the bi-stream network to preserve clear object boundaries.
To demonstrate the advantages of our method, we conducted massive quantitative comparisons against 16 state-of-the-art methods over 5 frequently used datasets. In summary, the contributions of this paper can be summarized as follows:
\begin{itemize}
\vspace{0.2cm}
\item  We provide a deeper insight into the interrelationship between the performance and training dataset;

\vspace{0.2cm}
\item  We propose a novel way to automatically construct small-scale training set MD4K from the off-the-shelf training datasets and our proposed MD4K  boost the state-of-the-art models performance consistently;

\vspace{0.2cm}
\item We design a bi-stream network with a novel gate control unit and multi-layer attention module. It can better mine the complementary information encoded in different network structures and help the network take full advantage of the proposed small dataset;

\vspace{0.2cm}
\item
Experimental results demonstrate that the proposed model achieves the state-of-the-art performance on five datasets in terms of six metrics, which proves the effectiveness and superiority of the proposed method.
\end{itemize}

\vspace{0.2cm}
\section{Related Works}
To simulate the human visual attention, early image SOD methods mainly focus on the hand-crafted visual features, cues and priors such as center prior~\cite{liu2011learning,judd2009learning}, background cues~\cite{li2013saliency,wei2012geodesic}, regional contrast~\cite{cheng2015global} and other kinds of relevant low-level visual cues~\cite{8798692,7776942}.
Due to the space limitation, we only concentrate on deep learning based SOD models here.


%
%
%
%
%
%
%
%
%
%

\begin{figure*}[t]
	\centering
	\includegraphics[width=\linewidth]{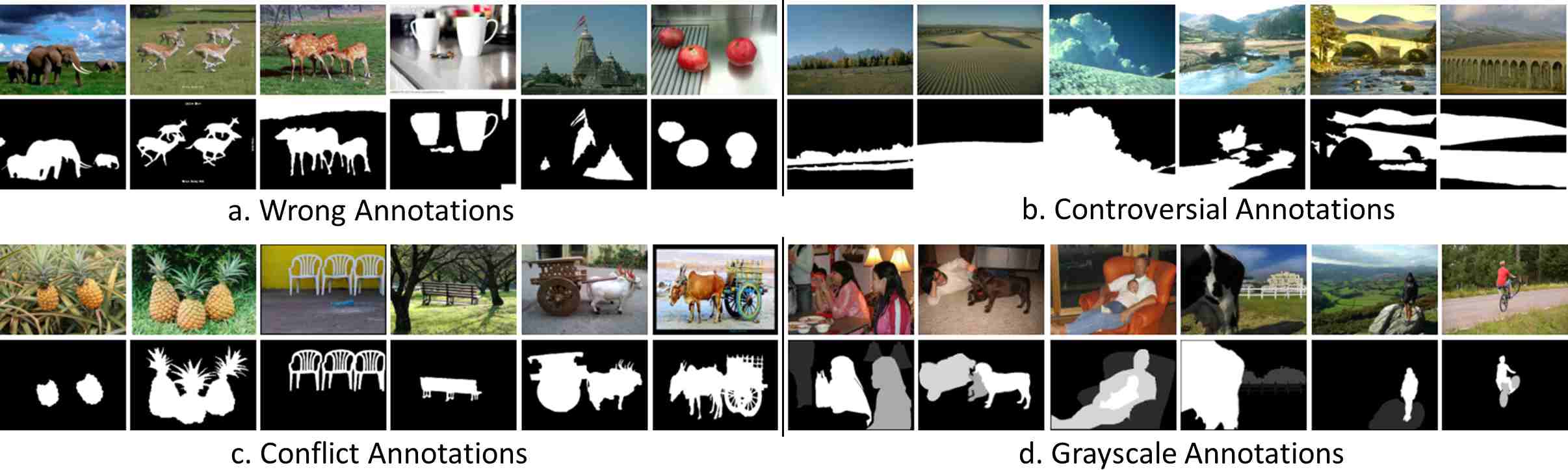}
	\caption{Examples of those inappropriate human annotations in the current SOD benchmarks, which are quite normal and can be divided into the above mentioned four groups, accordingly.}
	\label{fig3}
\end{figure*}

\subsection{Single-stream Model}
Generally, the deep network performance can be boosted significantly by aggregating the multi-level and multi-scale deep features between different layers.
As one of the most representatives, Hou~\textit{et al}.~\cite{DSS} proposed a top-down model to integrate both high-level and low-level features, achieving much improved SOD performance.
Following this rationale, various feature aggregation schemes~\cite{SRM,Amulet,RADF,DGRL,7895181,BMP,liu2016dhsnet,wang2019an,zhao2019egnet} were proposed latter.
Zhang \textit{et al.} \cite{Amulet} first integrate multi-level feature maps into multiple resolutions, which simultaneously incorporate semantic information and spatial details. Then this work predicts the saliency map in each resolution and fuses them to generate the final saliency map.
Liu and Han \cite{liu2016dhsnet} first make a coarse global prediction, and then hierarchically and progressively refine the details of saliency maps step by step via integrating local context information.
Zhang~\textit{et al.}~\cite{BMP} proposed a bi-directional structure with a gate unit to control information flow between multi-level features.
Wang \textit{et al.} \cite{wang2019an} proposed a novel schema that integrates both top-down and bottom-up saliency inference in an iterative and cooperative manner.
Zhao \textit{et al.} \cite{zhao2019egnet} present an edge guidance network for salient object detection with three steps to simultaneously model these two kinds of complementary information in a single network.
Wang \textit{et al.} \cite{wang2019inferring}  build a novel attentive saliency network that learns to detect salient objects from fixations, which narrows the gap between salient object detection and fixation prediction.
Compared to the gate setting proposed in~\cite{BMP}, the major highlight of our gate control unit is that it has achieved the full interactions between two different sub-networks by integrating complementary semantical information mutually.
Additionally, our gate control unit can well preserve the non-linear capabilities, enabling faster convergence and speed up training, more details can be found in Sec.~\ref{gate_control_unit}.

\subsection{Two-stream Network}

In recent years, the two-stream network has achieved much attention due to its effectiveness to many computer vision applications, including visual question answering \cite{saito2017dualnet}, image recognition \cite{hou2017dualnet,lin2015bilinear}, salient object detection \cite{zhao2015saliency, zhang2019capsal, zhou2020interactive}. Saito \textit{et al.} \cite{saito2017dualnet} propose to use different kinds of networks to extract image features in order to fully take advantage of different information present in different kinds of network structures. Lin \textit{et al.} \cite{lin2015bilinear} propose bilinear models, a recognition architecture that consists of two feature extractors whose outputs are multiplied using outer product at each location of the image and pooled to obtain an image descriptor. Hou \textit{et al.} \cite{hou2017dualnet} present a framework named DualNet to effectively learn more accurate representation for image recognition. The core idea of DualNet is to coordinate two parallel DCNNs to learn features complementary to each other, and thus richer features can be extracted from the raw images. Besides, recently, two-stream network structure also is adopted by SOD.  Zhao~\textit{et al}.\cite{zhao2015saliency} proposed a multi-context deep learning framework, in which the global context and local context are combined in a unified deep learning framework.
 Zhang \textit{et al.} \cite{zhang2019capsal} propose a new deep neural network model named CapSal which consists of two sub-networks, to leverage the captioning information together with the local and global visual contexts for predicting salient regions. Zhou \textit{et al.} \cite{zhou2020interactive}  propose a lightweight two-stream model that uses two branches to learn the representations of salient regions and their contours respectively.  All of the previous works mentioned above have demonstrated the effectiveness of the two-stream network and potentially prove this idea is good.
 Inspired but different from previous works, we propose a novel bi-stream network, consisting of two sub-branches with different network structures, which is aim to  take advantage of rich semantic information present in the proposed MD4K datasets.

\subsection{Attention Mechanism}
The ``attention mechanism'' has been widely used to boost the state-of-the-art methods performances~\cite{AFNet,PAGRN, PiCANet}, here, we will introduce several most representative approaches.
Inspired by human perception process, attention mechanism is introduced by using high-level information to efficiently guide bottom-up feedforward process, and it has achieved great success in a lot of tasks. In \cite{li2017instance-level, chen2016attention}, attention model was designed to weight multi-scale features. In \cite{wang2017residual}, residual attention module was stacked to generate deep attention aware features for image classification. In \cite{hu2019squeeze-and-excitation}, channel attention was first proposed to select representative channels. After that, it has been widely applied in various tasks including semantic segmentation \cite{yu2018learning}, image deraining \cite{li2018recurrent}, image super-resolution \cite{zhang2018image}. Recently, Zhang~\textit{et al}.~\cite{PAGRN} introduced both the spatial-wise and channel-wise  attention to the SOD task. Wang~\textit{et al}. \cite{wang2019salient} devise an essential pyramid attention structure for salient object detection, which enables the network to concentrate more on salient regions while exploiting multi-scale saliency information.
Liu~\textit{et al}.~\cite{PiCANet} proposed a pixel-wise contextual attention mechanism to selectively integrate the global contexts into the local ones. In \cite{RANet20}, a novel reverse attention block was designed to highlight the prediction of the missing salient object  and guide side-output residual learning. In contrast, our novel multi-layer attention module aims to transfer the high-level localization information to the shallower layers, shrinking the given problem domain effectively.

\subsection{The Major Highlights of Our Method}
In sharp contrast to the previous works which merely focus on the elegant network designs, our research will inspire the SOD community to pay more attention to the training data, despite in its early stage, new state-of-the-art performance can be easily reached. The proposed bi-stream network, which is well designed for the proposed small MD4K dataset, aims to  take advantage of rich semantic information present in the proposed MD4K datasets. To our best knowledge, this is the first attempt to use a ``wider'' model with a small-scale training set yet outperform previous models which are trained on large-scale training sets.

\section{A Small-scale Training Set}
\label{data analysis}
Given a SOD deep model, its performance usually relies on two factors: 1) the specific training dataset and 2) the number of training data.
Previous works \cite{wang2019salient1,fan2018SOC} have discussed that the selected training dataset influences model performance. In this section, we provide a further and detailed discussion about the interrelationship between these factors and the network performances.

\begin{figure}[!t]
\centering
\includegraphics[width=\linewidth]{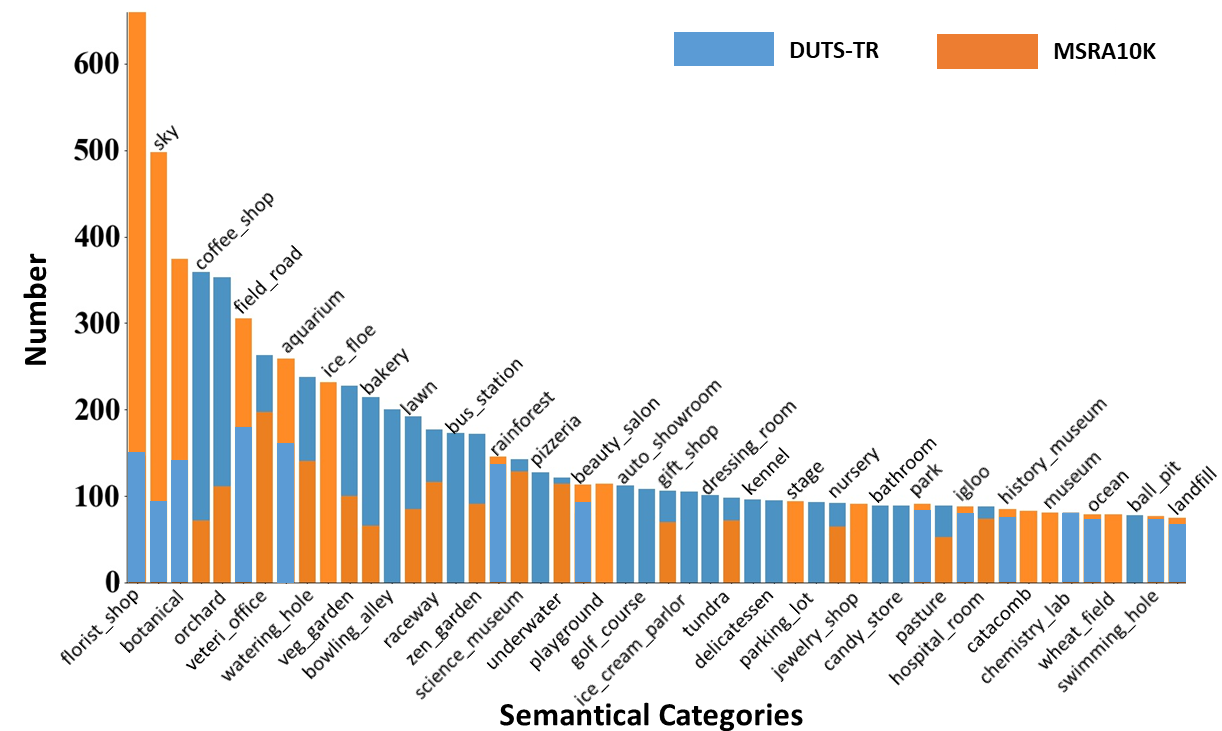}
\caption{The semantical category distributions (classified by~\cite{zhou2017places}) of the MSRA10K and the DUTS-TR datasets, indicating a strong semantical complementary status between these two datasets. We only demonstrate the top-50 categories due to the space limitation.}
\label{data_analysis_fig1}
\end{figure}

\subsection{Do We Really Need a Large-scale Training Data? }
Previous networks adopting complex network structures usually require a large-scale training data to reach their best performance.
This motivates us to consider a basic problem regarding the SOD task, i.e., will continually increasing the training data size be possible to achieve a steady SOD performance improvement?
To clarify this issue, we have trained the 2 state-of-the-art SOD methods, including the CPD19~\cite{CPD} and the PoolNet19~\cite{PoolNet}.
We first train the target SOD model on the whole DUTS-TR (10K) dataset and train the target model again using the DUTS-TR (9K) dataset which randomly removes 1000 images from the former training set, and repeating the above procedure.

The relationship between the overall performance and the training data number can be observed in Fig.~\ref{data_analysis_fig2}.
 As we can see, when training data increase to 2K, the performances have a significant improvement. However, with the training data continue growing, the performance is not always positively correlated with the amount of training data even get worse. Moreover, the performance trained on the whole DUTS-TR dataset is not the optimal result. Specifically, in terms of weighted F-measure, the performance of CPD19 on DUT-OMRON has been improved by about 12.5 $\%$ after training data increase to 2K. However, when the training data increased to 3K, the performance yet fell by 3.2 $\%$. The optimal performance is obtained when the training data equal to 6K instead of the whole DUTS-TR datasets. Similar conclusion can be obtained in other datasets or metrics.

 The primary reasons can be attributed to two-fold: 1) The unbalanced semantical categories in the original large-scale training set. For instance, by using the semantical labeling tool \cite{zhou2017places}, there are 351 images in the DUTS-TR dataset that are marked with the ``coffee shop'' semantical label, while the scenes of labeled with ``campus' is less than 10. And, the considerable redundant semantic scenes have less substantial help to improve performance. Moreover, previous works \cite{lake2011one,snell2017prototypical} have already demonstrated that CNN based model can able to understand new concepts given just a few examples. 2) There exists a considerable amount of bias annotations in the DUTS-TR training set, and such bias annotations even worse the overall performance as proofed in Fig.~\ref{data_analysis_fig2}.
In Fig. \ref{fig3}, we present several typical inappropriate human annotations, which motivates us to build a more clean training dataset to improve the SOD performance further.


\begin{center}
\begin{table}[!t]

\setlength{\abovecaptionskip}{10pt}
\setlength{\belowcaptionskip}{-10pt}

\scriptsize
\resizebox{1\textwidth}{!}{
\renewcommand\arraystretch{1.1}
\begin{tabular}{|r|cc|cc|cc|cc|}
\hline
\multirow{2}{*}{Method}
& \multicolumn{2}{c|}{DUT-OMRON}
& \multicolumn{2}{c|}{DUTS-TE}
& \multicolumn{2}{c|}{ECSSD}
\\
\cline{2-7}
&  avg$F_\beta$ & MAE  &avg$F_\beta$ & MAE  &  avg$F_\beta$ & MAE    \\

\hline
PoolNet19(MK)  & {0.702} &  {0.069}&  0.726 & 0.068 & \textbf{\textcolor{black}{0.888}}&  0.050  \\

PoolNet19(DTS)    &  \textbf{\textcolor{black}{0.738}}& \textbf{\textcolor{black}{0.055}}&  \textbf{\textcolor{black}{0.781}} &\textbf{\textcolor{black}{0.040}} &  0.880& \textbf{\textcolor{black}{0.049}}  \\
\hline

CPD19(MK) & 0.716 &  0.073 & 0.732 & 0.068 & 0.882 & 0.050 \\

CPD19(DTS)& \textbf{\textcolor{black}{ 0.738}} &  \textbf{\textcolor{black}{0.056}} & \textbf{\textcolor{black}{0.784}} & \textbf{\textcolor{black}{0.044}} & \textbf{\textcolor{black}{0.880}} & \textbf{\textcolor{black}{0.037}}\\
\hline

AFNet19(MK) & \textbf{\textcolor{black}{0.734}} &   \textbf{\textcolor{black}{0.053}} & \textbf{\textcolor{black}{0.786}} &  \textbf{\textcolor{black}{0.042}} & \textbf{\textcolor{black}{0.877}} &  \textbf{\textcolor{black}{0.040 }}  \\
AFNet19(DTS) &  0.729 &  0.057 & 0.772 & 0.046 & 0.871&  0.042 \\

\hline
\end{tabular}
}
\caption{Comparisons of the 3 state-of-the-art models trained on different datasets, where MK and DTS stand for MSRA10K and DUTS-TR respectively, and we use {\textcolor{black}{\textbf{bold}}} to emphasize better results.}
\label{data_analysis_tab1}
\end{table}
\end{center}

\subsection{Which Training Set Should be Selected?}
We noticed that most of the state-of-the-art models are typically trained on either the MSRA10K or the DUTS-TR dataset, then be evaluated on the others.
However, this training strategy suffers from serious limitations; i.e., the data distribution inconsistency between training and testing datasets may easily lead to the ``domain-shift'' problem.
For example, the images in the widely used training set MSRA10K are attributed as high contrast, center-surround, simple background, and containing single salient objects only.
However, the images in commonly used testing set HKU-IS~\cite{zhao2015saliency} are attributed as low contrast, relative complex background, and usually containing multiple salient objects. Although the DUTS-TR dataset is complex, it introduces additional challenging problems such as non-inconsistent saliency ground-truth and controversial annotation. This motivates us to combine their advantages of MSRA10K and DUTS-TR datasets.

Actually, as the commonly used training sets, the MSRA10K and the DUTS-TR datasets are complementary in general.
To back our claim, we have tested the 3 state-of-the-art SOD models in Table~{\ref{results_tab1}}, in which these models are trained on MSRA10K and DUTS-TR datasets respectively and then tested on others.
As shown in Table {\ref{data_analysis_tab1}}, we may reach to a sub-optimal training performance if we only use either the MSRA10K or the DUTS-TR training set.
Also, we have demonstrated the semantical category distribution of the MSRA10K and the DUTS-TR datasets in Fig. {\ref{data_analysis_fig1}}, which shows a large semantical variance between these two datasets, showing their semantical complementary.

 On the other hand, previous works \cite{wang2017learning, zhang2019capsal, zeng2019multi-source,wang2019robust,hsu2017weakly} have already demonstrated that semantic information, especially in cluttered scenes, is beneficial to the SOD task. Wang \textit{et al.} \cite{wang2019robust} propose a novel end-to-end deep learning approach for robust co-saliency detection by simultaneously learning high-level group-wise semantic representation as well as deep visual features of a given image group.
To address accurately detect salient objects in cluttered scenes,  the author of \cite{zhang2019capsal} argues that the model needs to learn discriminative semantic features for salient objects, such as object categories, attributes and the semantic context. Therefore, it is necessary to build a semantical category balanced training dataset to further improve the SOD performance.

\begin{figure*}[!t]

\centering
\includegraphics[width=\linewidth]{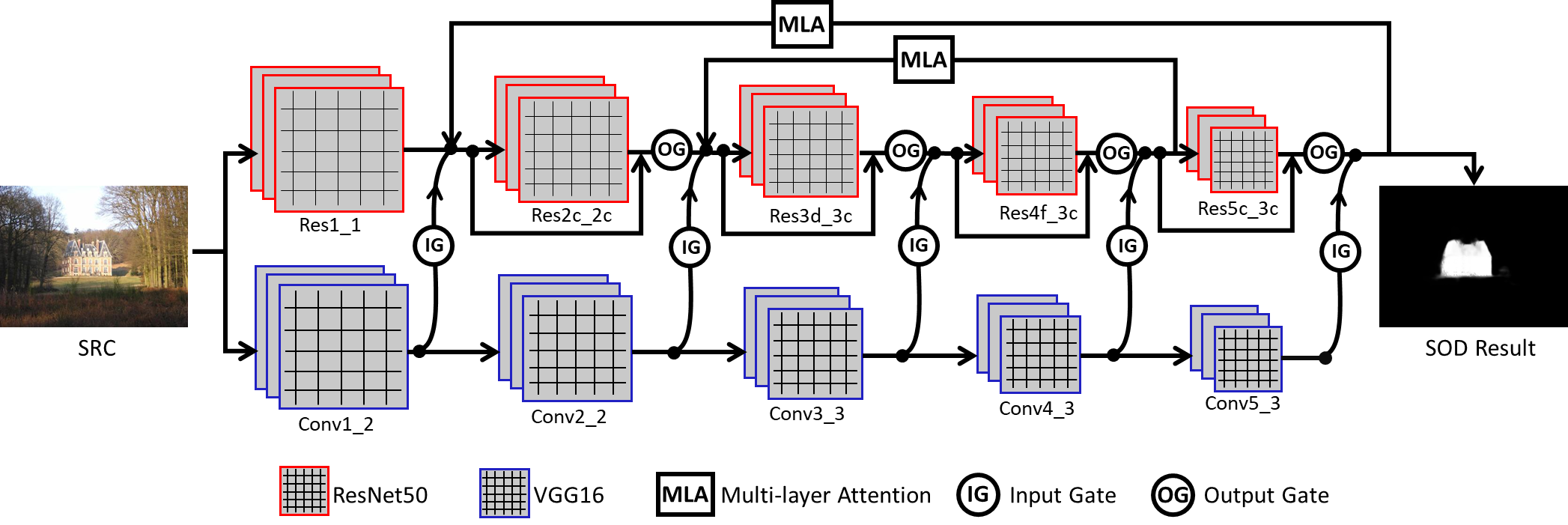}

\caption{The detailed architecture of the proposed bi-stream network. Our bi-stream network is developed on the commonly used ResNet50 and VGG16, using both the newly designed gate control unit (Sec.~\ref{gate_control_unit}) and the scaling-free multi-layer attention (Sec.~\ref{multi-layer_attention}) to achieve the complementary status between two parallel sub-branches, which is also capable of taking full advantage of the multi-level deep features as well.
 }
\label{method_fig1}
\end{figure*}

\subsection{Our Novel Training Dataset Construction}
In this section, we build a small, GT bias-free and semantical category balanced training dataset from the MSRA10K and the DUTS-TR datasets, namely ``MD4K''.
The motivation can be summarized as the following 4 aspects:
1) According to our experiment, the performance is not always positively correlated with the amount of training data;
2) The off-the-shelf SOD models can not achieve the optimal performances by using single training set solely;
3) Existing training sets contain massive dirty and unbalanced data;
4) The MSRA10K and DUTS-TR datasets are complementary as mentioned before.

We first divided MSRA10K and DUTS-TR datasets into 267 categories utilizing the off-the-shelf scene classification algorithm \cite{zhou2017places}.
Then, we manually remove all those dirty data, thus there are 9012 left in the MSRA10K dataset and 9215 images left in the DUTS-TR dataset.
Interestingly, we found that the semantical category distribution of the above 18K images obeys the Pareto Principle, i.e., $20\%$ scene categories are account for $80\%$ of the total.
Specifically, the top-50 scene categories of MSRA10K account for $71.23\%$ of the whole MSRA10K dataset, and such percentage is $74.13\%$  in the DUTS-TR dataset.
To balance the semantical categories, we randomly select 40 images for each of the top-50 scene categories and then choose 20 images for each of the remaining 217 scene categories.
In this way, we finally obtain a small-scale training set, containing 4172 images with total 267 semantical categories.
The reason we choose 4172 images is that we attempt to find a balance between training size and performance, and the performance trained on a different number of data is shown in Table \ref{MD4K_number_size_analysis}.
 According to the experimental results, the training set with 4172 images can achieve better performance than DUTS-TR meanwhile decrease the training data number significantly.

The significance of the proposed MD4K dataset can be summarized as follows:
1) The proposed MD4K can alleviate the demands for large-scale training data;
2) Our proposed MD4K boost the state-of-the- art models performance consistently;
3) Our MD4K may inspire other researchers about how to build a training set.

\begin{table}[htbp]
\begin{center}
\setlength{\abovecaptionskip}{0pt}%
\setlength{\belowcaptionskip}{5pt}%
\renewcommand\arraystretch{1.5}
\Large
\resizebox{1\textwidth}{!}{
\centering
\begin{tabular}{|r|>{\columncolor{mygray}}c|c|c|c|>{\columncolor{mygray}}c|c|c|}
\hline
\diagbox{Tested on}{Trained on}& DUTS-TR& MD1K & MD2K & MD3K & MD4K & MD5K & MD6K  \\
\hline
DUT-OMRON \cite{yang2013saliency} &0.835 & 0.715 & 0.794& 0.832 & 0.857 &0.864 & 0.866  \\
 \hline
 DUTS-TE \cite{wang2017learning} &0.879 & 0.774 & 0.829& 0.863   & 0.884 &0.893 &0.897   \\
 \hline
 ECSSD \cite{ecssd}&0.934 & 0.876  & 0.876 & 0.918    & 0.945 &0.947 &0.955   \\
 \hline
 HKU-IS \cite{zhao2015saliency}&0.933 & 0.864  & 0.885& 0.920    & 0.942 &0.948 &0.952   \\
\hline
PASCAL-S \cite{li2014secrets}&0.885 & 0.778  & 0.837& 0.864   & 0.886 & 0.895&0.897   \\
\hline

\end{tabular} }
\vspace{0.2cm}
\caption{Performance trained on a different number of MD4K data. For each dataset, we use the average max F-measure to evaluate their performance.}
\vspace{-0.6cm}
\label{MD4K_number_size_analysis}
\end{center}
\end{table}

\vspace{-0.2cm}
\section{Proposed Network}
\label{method}
So far, we have built a small-scale and high-quality training dataset which can consistently boost the state-of-the-art performances, see the quantitative proofs in Table~\ref{ablation_studies_tab2}.
To further improve, we propose a novel bi-stream network consisting of two feature backbones with different structures, aiming to sense complementary semantical information, taking full advantage of our semantical balanced small-scale training set.

\subsection{How to Fuse Bi-stream Networks}
\label{gate_control_unit}

In this section, we consider how to effectively fuse two different feature backbones, in which we attempt to use feature maps extracted from one sub-branch to benefit another one.
We shall provide some preliminaries regarding the conventional common threads here.

For simplicity, the function {\textit{f}}: \{$(\textbf{X}^R,\textbf{X}^V)\rightarrow\textbf{Y}$\} represents fusing two feature maps {$\textbf{X}^R$} and {$\textbf{X}^V$} to generate the output feature {$\textbf{Y}$}, where \{$\textbf{X}^R,\textbf{X}^V,\textbf{Y}\in \mathbb{R}^{H \times W \times C}$\}, $H,W,C$ denote the height, width and channels respectively.

\vspace{0.1cm}
\noindent 1) \textbf{Element-wise summation}, {$\textbf{Y}_{sum}$}, which calculates the sum of two features at the same locations ($w,h$) and channels ($c$):
\begin{equation}
\label{method_formulation_1}
\textbf{Y}_{sum}= \sum \limits_{c=1}^{C} \sum \limits_{w=1}^{W} \sum \limits_{h=1}^{H} (\textbf{X}^{R}_{h,w,c} + \textbf{X}^{V}_{h,w,c}),
\end{equation}

\noindent 2) \textbf{Element-wise maximum}, $\textbf{Y}_{max}$, which analogously takes the maximum of two input feature maps:
\begin{equation}
\label{method_formulation_2}
\textbf{Y}_{max}= \sum \limits_{c=1}^{C} \sum \limits_{w=1}^{W} \sum \limits_{h=1}^{H} max(\textbf{X}^{R}_{h,w,c}, \textbf{X}^{V}_{h,w,c}),
\end{equation}

\noindent 3) \textbf{Concatenation}, {$\textbf{Y}_{concat}$}, which stack the input feature maps channel-wisely:
\begin{equation}
\label{method_formulation_3}
\textbf{Y}_{concat} =Concat(\textbf{X}^{R}_{h,w,c} , \textbf{X}^{V}_{h,w,c}),
\end{equation}

\noindent 4) \textbf{Convolution}, {$\textbf{Y}_{conv}$}, which first employ the concatenation operation to obtain features {$ \textbf{Y}_{concat} \in \mathbb{R}^{H \times W \times {2C}}$} and then convolve it:
\begin{equation}
\label{method_formulation_4}
\textbf{Y}_{conv} =  \textbf{Y}_{concat} * \textbf{W} + \textbf{b},
\end{equation}
where $*$ denotes the convolution operation, {\textbf{W}} represents the convolution filters, and \textbf{b} denotes the bias parameters.


\begin{figure*}[!t]
\centering
\includegraphics[width=\linewidth]{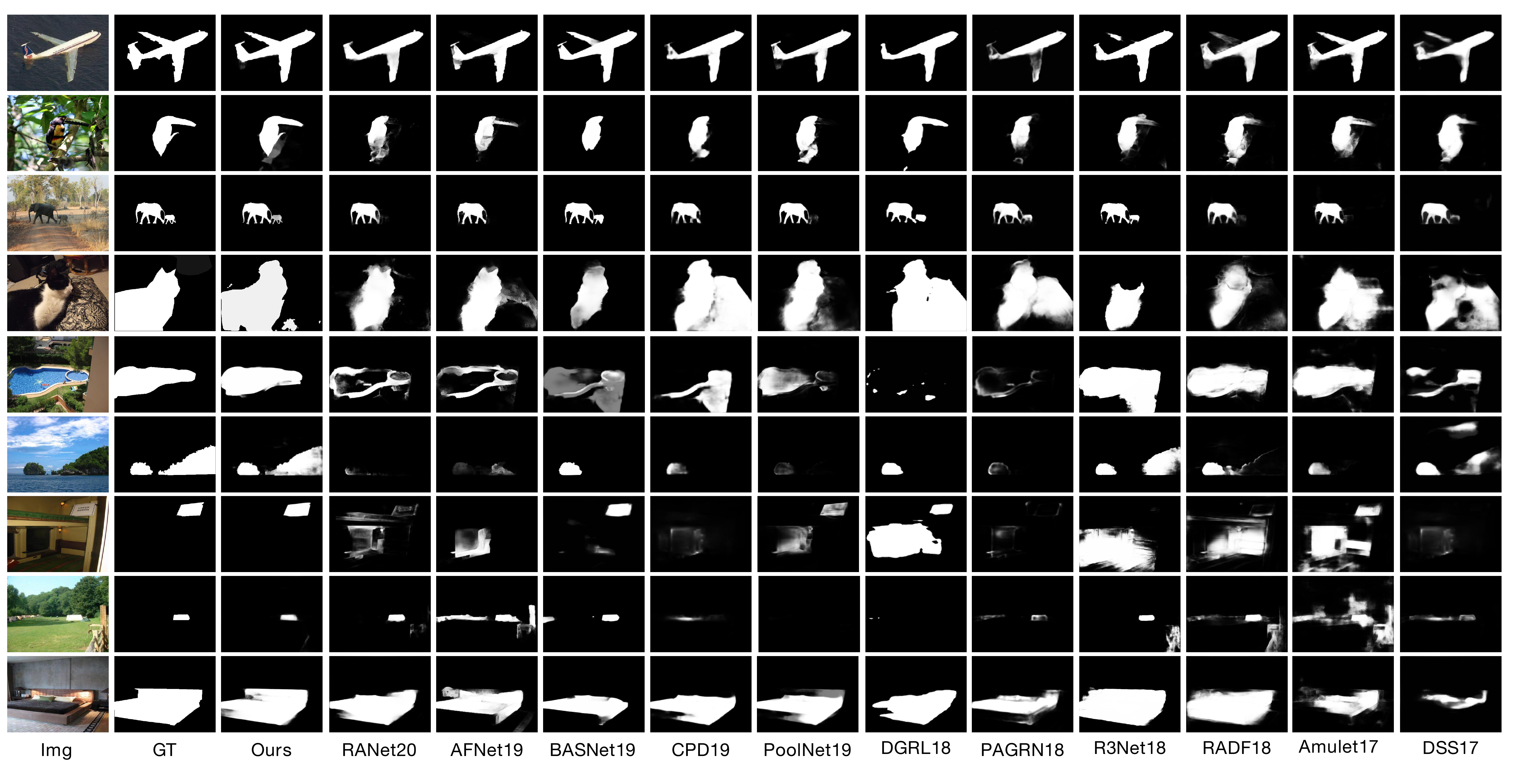}
\caption{Qualitative comparisons to the recent state-of-the-art models. Our approach can well locate the salient objects completely with sharp boundaries.} \vspace*{-0.4cm}
\label{result_maps1}
\end{figure*}

\subsection{Bi-stream Fusion via Gate Control Unit}
In general, all of the above-mentioned fusion operations directly fuse two input feature maps without considering the feature conflictions between different layers, which easily lead to the sub-optimal results, see the quantitative proofs in Table~\ref{ablation_studies_tab1}.
Inspired from the previous work~\cite{LSTM}, we propose a novel gate control unit, i.e., input gate and output gate, to control which information flows in the network, where the Fig. {\ref{method_fig1}} illustrates our novel network architecture.
In our method, the proposed input gate play a critical role in aggregating feature maps.
For clarity, let {$\textbf{X}^V$} = \{{$\textbf{X}^V_i$}, $i = 1,..., 5$\} denotes the feature maps for each convolutional blocks of the pre-trained VGG16 feature backbone.
Similarly, {$\textbf{X}^R$} represents the feature maps of the pre-trained ResNet50 backbone.

We introduce the dynamic thresholding in the proposed input gate, in which each side-output of VGG16 with a probability below the threshold will be suppressed.
Specifically, each side-output of VGG16 is a linear projection { $\textbf{X}^V_i*\textbf{W} +\textbf{b}$} modulated by the gates {${\sigma(\textbf{X}^V_i*\textbf{V}_{in} + \textbf{b}_{in})}$}.

In practice, the input gate will be element-wisely multiplied by the side-output feature matrix {$\textbf {X}^V_i *\textbf{ W}+\textbf{b}$}, controlling the interactions between the parallel sub-branches hierarchically.
Thus, the fused bi-stream feature maps ($\textbf{Y}_{conv}$) can be obtained by using the below operation.
\begin{equation} \label{method_formulation_5}
  \begin{split}
  \Theta(\textbf{X}^V_i)&=(\textbf{X}^V_i * \textbf{W} +\textbf{b}) \otimes {{\sigma(\textbf{X}^V_i *\textbf{ V}_{in} + \textbf{b}_{in})}},  \\
  \textbf{Y}_{conv} &= f\big(\textbf{X}^R_i,\Theta(\textbf{X}^V_i)\big),
  \end{split}
\end{equation}
where $\textbf{W}$, $\textbf{b}$, $\textbf{V}_{in}$, $\textbf{b}_{in}$ are learned parameters, $\sigma$ is the sigmoid function and $\otimes$ is the element-wise multiplication operation.

Moreover, previous SOD models directly propagate the feature maps from low-layer to high-layer without considering whether these features are beneficial to the SOD task.
In fact, only a small part of these features are useful, yet others may lead to even worse performance.
To solve this problem, we propose a multiplicative operation based ``output gate'' to suppress those distractions from the non-salient regions.
That is, given two consecutive layers, the feature responses in the precedent layer {$\sigma(\textbf{X}^R_{i-1} *\textbf{ V}_{out} + \textbf{b}_{out})$} will serve as the guidance for the next layer {$\textbf{ {X}}^R_i $} $ (i\in \{2,3,4,5\})$ to adaptively control which data flow should be propagated automatically, and this procedure can be formulated as Eq.~\ref{method_formulation_6}.
\begin{equation} \label{method_formulation_6}
\tau(\textbf{X}^R_i,\textbf{X}^R_{i-1}) =\textbf{X}^R_i  \otimes { \sigma(\textbf{X}^R_{i-1} *\textbf{ V}_{out} + \textbf{b}_{out})},
\end{equation}
where $\textbf{V}_{out}$, $\textbf{b}_{out}$ is the learned weights and biases.
In this way, the salient regions which have high responses will be enhanced while the background regions will be suppressed in subsequent layers.
Consequently, our gate control unit constantly boost the conventional fusion performances, see the quantitative proofs in Table~\ref{ablation_studies_tab1}.

\vspace{0.2cm}
\noindent\textbf{Differences to the LSTM}.\\
\vspace{0.2cm}
The gradient in original LSTM~\cite{dauphin2015predicting} can be expressed as:
\begin{equation} \label{method_formulation_7}
\begin{split}
\nabla\big(tanh(\textbf{X})  \otimes \sigma(\textbf{X})\big) =  \sigma'(\textbf{X})\nabla\textbf{X} \otimes tanh(\textbf{X}) \\+   tanh'(\textbf{X}) \nabla\textbf{X} \otimes \sigma(\textbf{X}).
\end{split}
\end{equation}
Notice that such gradient will gradually get vanished due to the down-scaling factor $tanh'(\textbf{X})$ and $\sigma'(\textbf{X})$.
In sharp contrast, the gradient of our gate mechanism has a directional path $ \nabla\textbf{X} \otimes \sigma(\textbf{X})$ without using any down-scaling operations for the activated gating units in $\sigma(\textbf{X})$ as Eq.~\ref{method_formulation_8}.
\begin{equation} \label{method_formulation_8}
\begin{split}
\nabla\big(\sigma(\textbf{X})\otimes\textbf{X}\big) =  \nabla\textbf{X} \otimes \sigma(\textbf{X})  +  \sigma'(\textbf{X})\nabla\textbf{X} \otimes \textbf{X},
\end{split}
\end{equation}
Thus, the proposed gate control unit outperforms the LSTM significantly, see the quantitative proofs in the Table~\ref{ablation_studies_tab1}, i.e., ``Conv w/ GCN (Ours)'' vs. ``Conv w/ GCN (LSTM)''.

%
%

\begin{center}
\begin{table*}[!t]
\setlength{\abovecaptionskip}{0pt}%
\setlength{\belowcaptionskip}{5pt}%

\small{
\linespread{2}
\renewcommand\arraystretch{1.1}

\resizebox{1\textwidth}{!}{

\begin{tabular}{|r|c|cc|cc|cc|cc|cc|cc|cc|}
\hline

\multirow{2}{*}{Method}
&\multirow{2}{*}{Backbone}
& \multicolumn{2}{c|}{Training}
& \multicolumn{2}{c|}{DUT-OMRON}
& \multicolumn{2}{c|}{DUTS-TE}
& \multicolumn{2}{c|}{ECSSD}
& \multicolumn{2}{c|}{HKU-IS}
& \multicolumn{2}{c|}{PASCAL-S}
\\
\cline{3-14}
& & Images& Dataset &max$F_\beta\uparrow$  & MAE$\downarrow$    &max$F_\beta\uparrow$  & MAE$\downarrow$   &max$F_\beta\uparrow$& MAE$\downarrow$  &max$F_\beta\uparrow$  & MAE$\downarrow$  &max$F_\beta\uparrow$ & MAE$\downarrow$ \\
\hline

\textbf{Ours} &ResNet50+VGG16& 4172& MD4K&{\textcolor{red}{0.857}}&{\textcolor{red}{0.044}} & {\textcolor{red}{0.884}} &  {\textcolor{red}{0.038}} & {\textcolor{red}{0.945}} &\textcolor{red}{0.036} & {\textcolor{red}{0.942}} & {\textcolor{SeaGreen3}{0.031}} & {\textcolor{red}{0.886}}  &  {\textcolor{blue}{0.082}} \\

{Ours} & ResNet50+VGG16&10553& DTS& \textcolor{SeaGreen3}{0.835}& \textcolor{SeaGreen3}{0.046}& \textcolor{SeaGreen3}{0.879}&\textcolor{blue}{0.041}&{0.934}&  \textcolor{blue}{0.039} &\textcolor{SeaGreen3}{0.933}& {0.033} &\textcolor{SeaGreen3}{0.885} &{0.089} \\

{Ours} & ResNet50+VGG16&10000& MK&  0.828 & \textcolor{blue}{0.047}& 0.863 & 0.044 & 0.931 & 0.042 &0.917 &0.035 & 0.857 & 0.088\\

{Ours} &ResNet50+ResNet50& 4172& MD4K&{\textcolor{blue}{0.833}} &{\textcolor{SeaGreen3}{0.046}} & {\textcolor{black}{0.855}} &  {\textcolor{blue}{0.041}} & {\textcolor{black}{0.921}} &{\textcolor{black}{0.043}} & {\textcolor{black}{0.916}} & {\textcolor{black}{0.037}} & {\textcolor{black}{0.853}} &   {\textcolor{black}{0.087}} \\

{Ours} &VGG16+VGG16& 4172& MD4K&{\textcolor{black}{0.826}} &{{0.049}} & {\textcolor{black}{0.849}} &  {\textcolor{black}{0.047}} & {\textcolor{black}{0.924}}&{\textcolor{black}{0.042}} & {\textcolor{black}{0.918}}& {{0.033}} & {\textcolor{black}{0.844}} &   {\textcolor{black}{0.092}} \\

{RANet20}\cite{RANet20}& VGG16&10553& DTS & 0.799	&	0.058&	\textcolor{blue}{0.874}	&	0.044	&	\textcolor{blue}{0.941}	&	0.042 &{0.928}	&	0.036&	{0.866}		&\textcolor{SeaGreen3}{0.078}\\

{{R$^2$Net20}}\cite{R2Net20} &VGG16&10553 &DTS & 0.793	&	0.061 & 	0.855	&	0.050&  	\textcolor{blue}{0.935}	&	0.044&	0.921	&	\textcolor{red}{0.030} & 	0.864&		\textcolor{red}{0.075}\\

{MRNet20}\cite{MRNet20} & ResNet50 &10553&DTS& 0.731	&	0.062&	0.792	&	0.048&	0.904	&0.048&0.891	&	0.039&	 	0.818	&	\textcolor{red}{0.075}\\

\textbf{CPD19}$^*$\cite{CPD}   & ResNet50& 4172& MD4K& \textbf{\textcolor{black}0.762}& \textbf{0.052} &\textbf{\textcolor{black}0.850} & \textbf{\textcolor{SeaGreen3}{0.040}} & \textbf{{0.934}}& \textbf{\textcolor{SeaGreen3}{0.037}} & \textbf{\textcolor{black}0.915}&\textbf{\textcolor{blue}{0.032}}  &\textbf{\textcolor{black}0.846}& \textbf{0.090}\\

CPD19\cite{CPD}  & ResNet50& 10553& DTS& {0.754}  &{0.056} & {0.841} &  {0.044} & {0.926}  &   \textcolor{SeaGreen3}{0.037} & {0.911}  & {0.034} & {0.843}  &  {0.092}\\

\textbf{PoolNet19}$^*$\cite{PoolNet} &   ResNet50&4172& MD4K&\textbf{\textcolor{black} 0.767} &  \textbf{0.051} &\textbf{{0.863}} &   \textbf{0.042} & \textbf{\textcolor{black}0.931} & \textbf{0.040} &\textbf{\textcolor{black}0.922} & \textbf{{0.033}} &\textbf{\textcolor{black}0.859} & \textbf{0.084}\\

PoolNet19\cite{PoolNet} &ResNet50&10553& DTS& {0.763} & {0.055}&{0.858}   & \textcolor{SeaGreen3}{0.040} & {0.920} & {0.042}& {0.917}   & {0.033} & {0.856}  &  0.093\\

\textbf{AFNet19}$^*$\cite{AFNet} & VGG16&4172 & MD4K&  \textbf{\textcolor{black}0.765} &  \textbf{0.054} &\textbf{\textcolor{black} 0.842} &   \textbf{0.044} &\textbf{{0.932}} &  \textbf{0.041} &\textbf{\textcolor{black}0.913}&  \textbf{0.034} &\textbf{\textcolor{black}0.854} &   \textbf{0.087}\\

AFNet19\cite{AFNet}  &VGG16&10553 & DTS& {0.759} &  {0.057} & {0.838}  &  0.046 &{0.924} & {0.042}& {0.910}  & 0.036 & {0.852}  & {0.089} \\

BASNet19\cite{BASNet19} &ResNet34&10553 & DTS&0.805 &0.057& 0.859&0.048 & \textcolor{SeaGreen3}{0.942}	&	\textcolor{SeaGreen3}{0.037}& \textcolor{SeaGreen3}{0.929}	&	\textcolor{blue}{0.032}&\textcolor{blue}{0.876}	&	0.092\\

MWS19\cite{MWS}  &DenseNet169& 310K&CO+DTS& {0.677} & 0.109 & {0.722} &  0.092 & {0.859}  &  0.096 & {0.835}  &   0.084 & {0.781} & 0.153\\

PAGRN18\cite{PAGRN} &VGG19&10553&DTS& {0.707}  & 0.071& {0.818}  & {0.056}& {0.904}  &0.061 & {0.897} & 0.048 &{0.817} &  0.120 \\

DGRL18\cite{DGRL} &ResNet50&10553&DTS& {0.739} & {0.062} & {0.806} & {0.051}& {0.914}  &{0.049}& {0.900} & {0.036} & {0.856} &   {0.085}\\

RADF18\cite{RADF}   &VGG16&10000&MK& {0.756} & {0.072} & {0.786} & {0.072} & {0.905}  &  0.060 & {0.895}  &  {0.050} & {0.817} & 0.123 \\

R$^3$Net18\cite{R3Net18} & ResNeXt&10000&MK& {0.460}    &  0.138 & {0.478} & 0.136 & {0.656}  & 0.161&  0.583 & 0.150 & {0.611}  &0.203\\

SRM17\cite{SRM}  &ResNet50&10553&DTS&{0.725}    & {0.069} & {0.799}  & {0.059} & {0.905}  &  {0.054}& {0.893}  &{0.046} & {0.812} &  {0.105}  \\

Amulet17\cite{Amulet} &VGG16&10000&MK& {0.715}    & 0.098 & {0.751} & 0.085 & {0.904} &  {0.059} & {0.884}  &0.052  & {0.836}  & 0.107 \\

UCF17\cite{UCF} &VGG16&10000&MK& {0.705} & 0.132 & {0.740}  & 0.118 & {0.897} &  0.078 & {0.871}& 0.074 & {0.820}  & 0.131 \\

DSS17\cite{DSS} &VGG16& 2500&MB&{0.681}  & 0.092 & {0.751} & 0.081 & {0.856} &  0.090 & {0.865} & 0.067 & {0.777} &  0.149  \\

\hline
 \end{tabular}  }}

\caption{The detailed quantitative comparisons between our method and 16 state-of-the-art models in \textbf{F-measure} and \textbf{MAE}.  Top three scores are denoted in \textcolor{red}{red}, \textcolor{SeaGreen3}{green} and \textcolor{blue}{blue}, respectively. \{MD4K, DTS, MK, MB, VOC, TH, CO\} are training datasets which respectively denote \{our small dataset, DUTS-TR, MSRA10K, MSRA-B, PASCAL VOC2007, THUS10K, Microsoft COCO\}. The symbol ``*'' indicates that the target models were trained on the MD4K dataset.}
\label{results_tab1}
\end{table*}
\end{center}



\begin{figure*}[htbp]
\centering

\subfigure{
\begin{minipage}[b]{0.202\linewidth}
\includegraphics[width=1\linewidth]{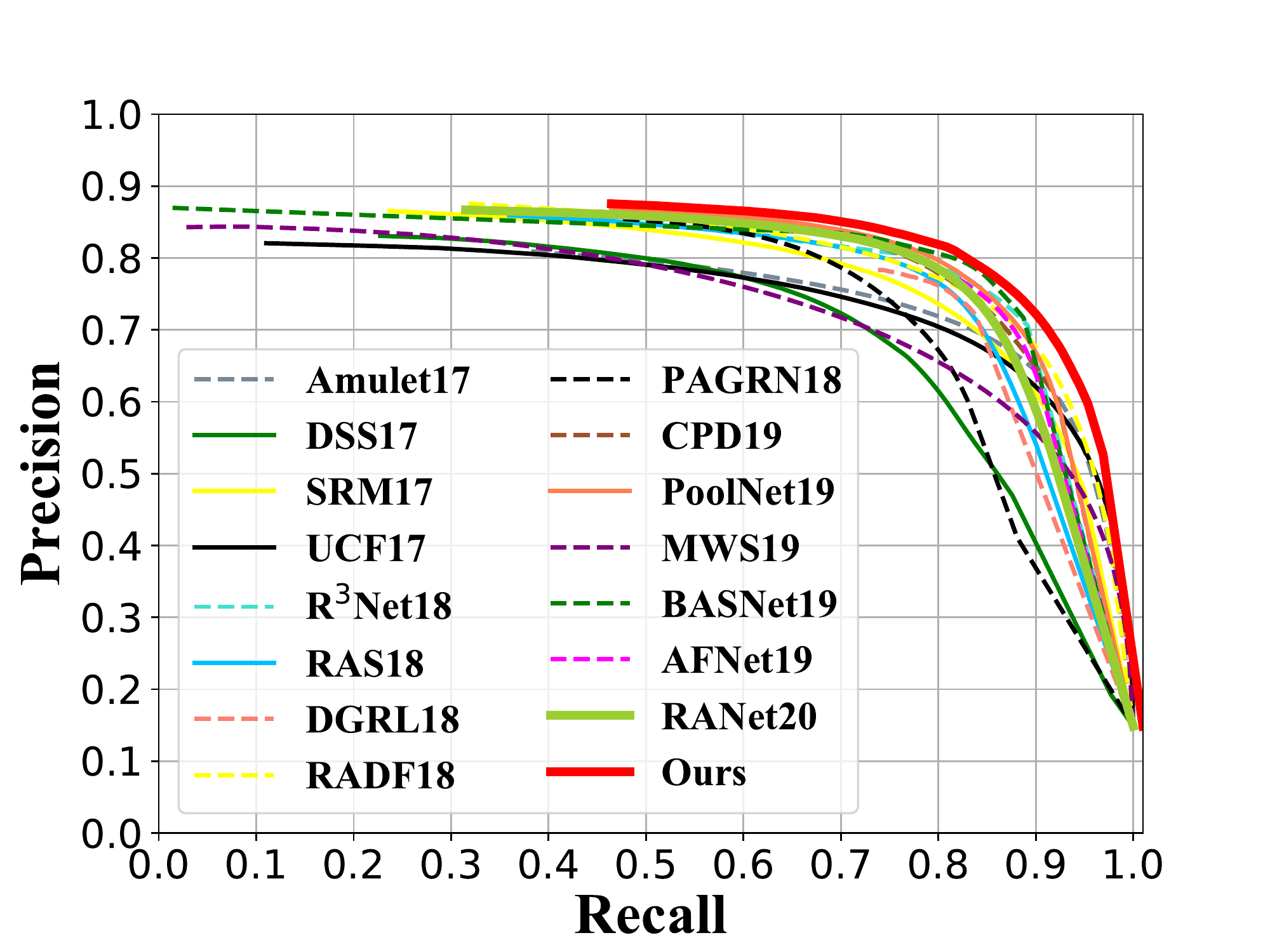}

\end{minipage}}\hspace{0.1pt}\hspace*{-1em}\vspace*{-0.3em}
\subfigure{
\begin{minipage}[b]{0.202\linewidth}
\includegraphics[width=1\linewidth]{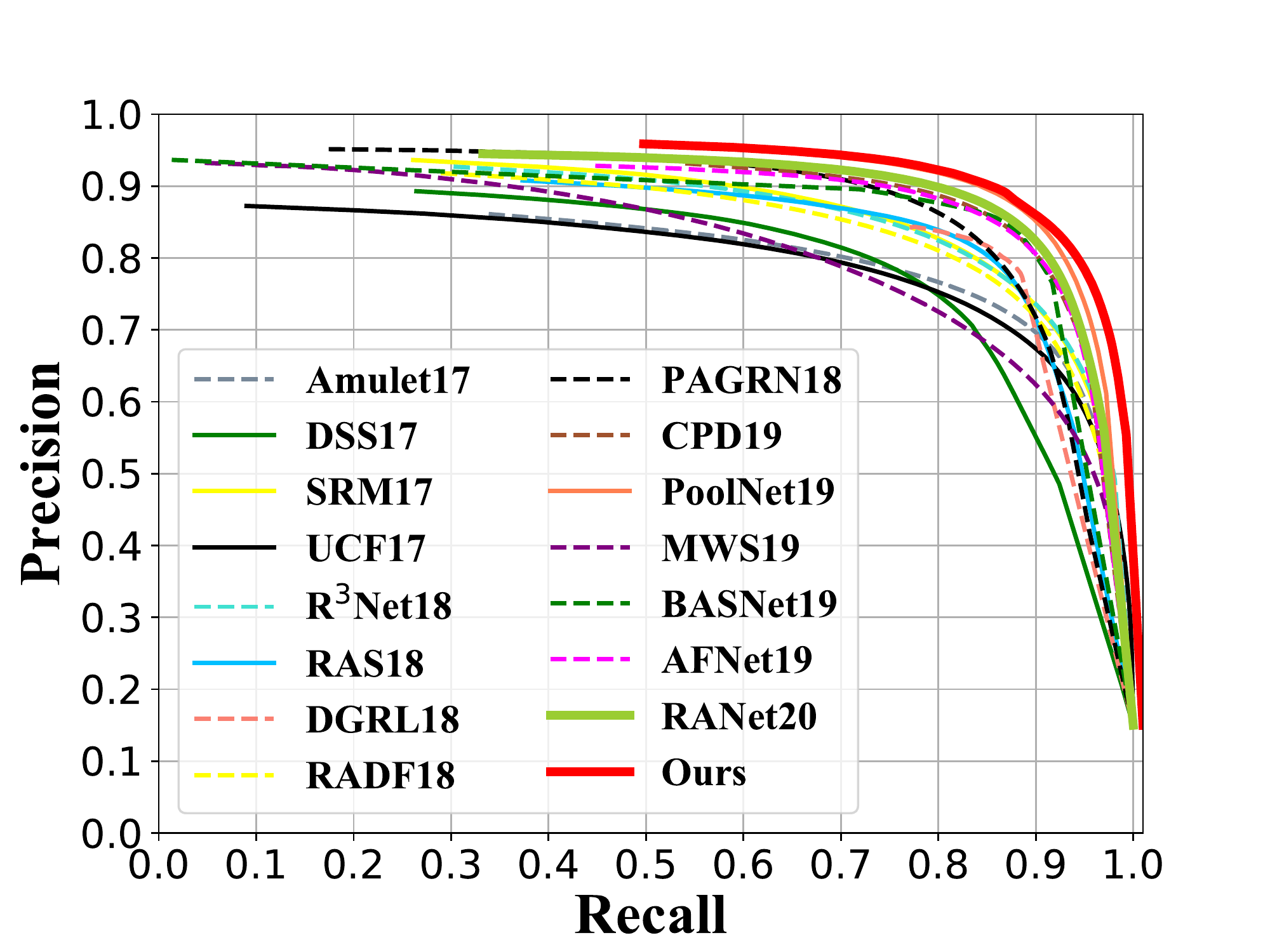}

\end{minipage}}\hspace{0.1pt}\hspace*{-1em}\vspace*{-0.3em}
\subfigure{
\begin{minipage}[b]{0.202\linewidth}\vspace*{-0.3em}
\includegraphics[width=1\linewidth]{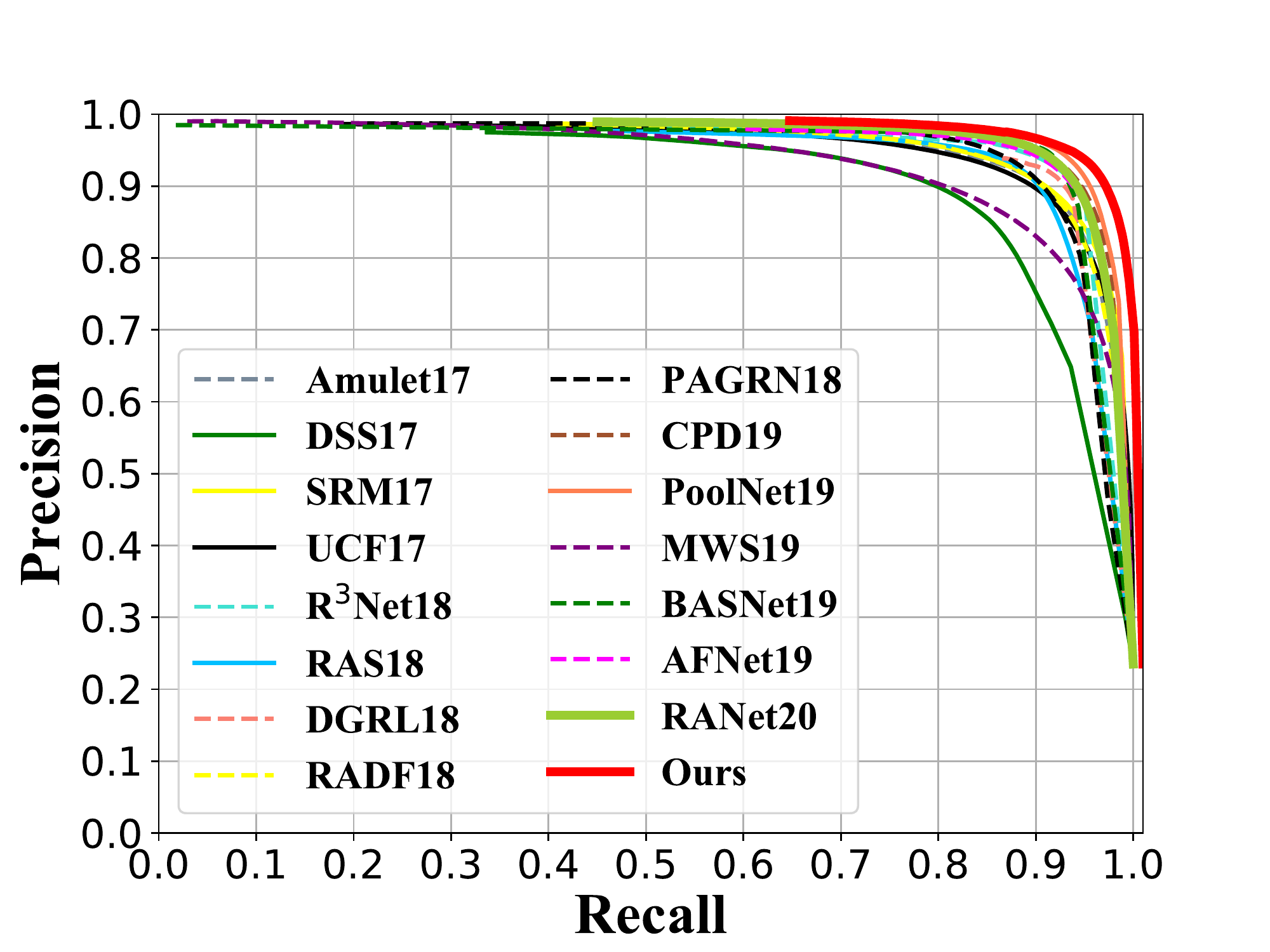}

\end{minipage}}\hspace{0.1pt}\hspace*{-1em}
\subfigure{
\begin{minipage}[b]{0.202\linewidth}\vspace*{-0.3em}
\includegraphics[width=1\linewidth]{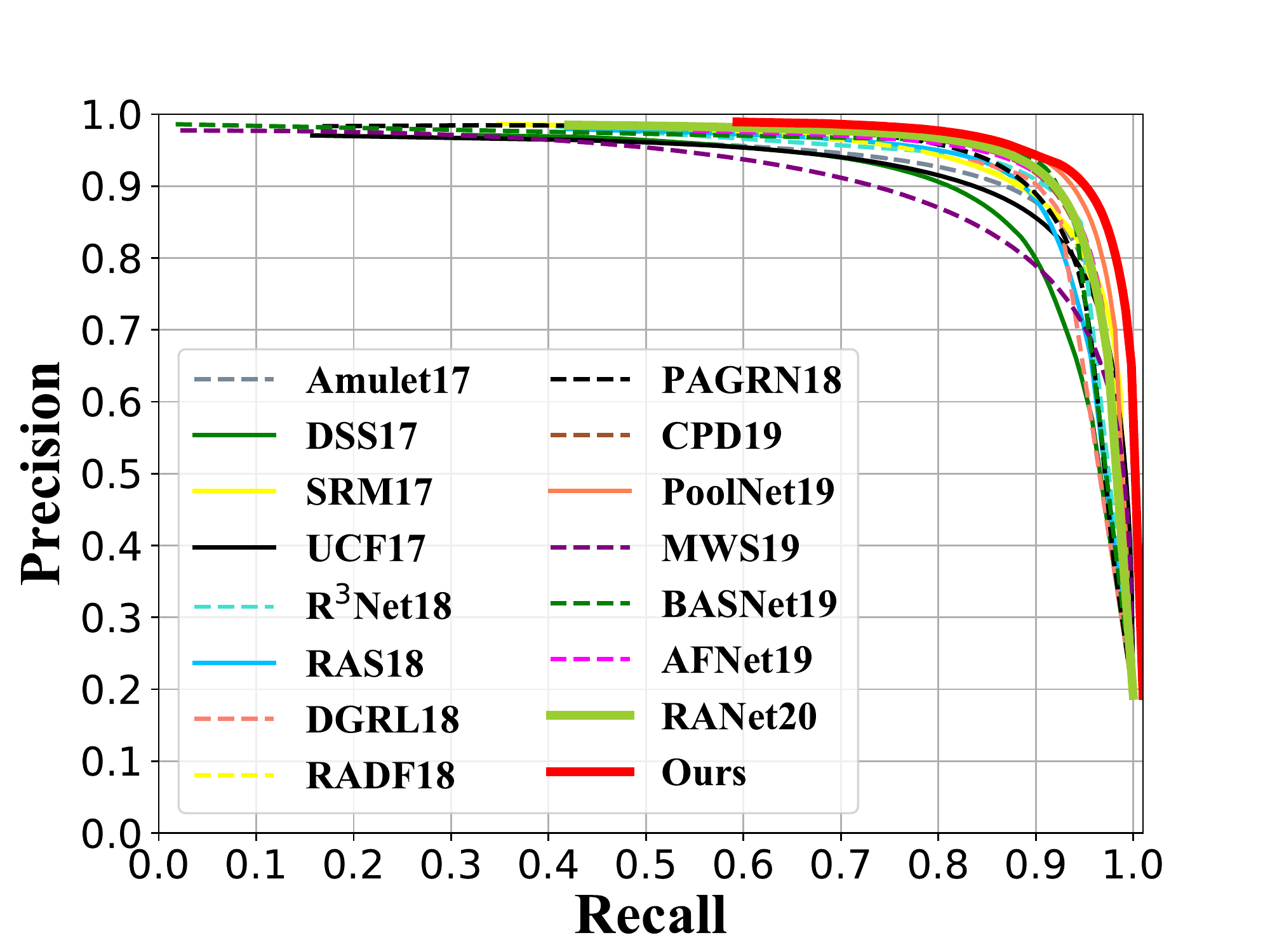}

\end{minipage}}\hspace{0.1pt}\hspace*{-1em}
\subfigure{
\begin{minipage}[b]{0.202\linewidth}\vspace*{-0.3em}
\includegraphics[width=1\linewidth]{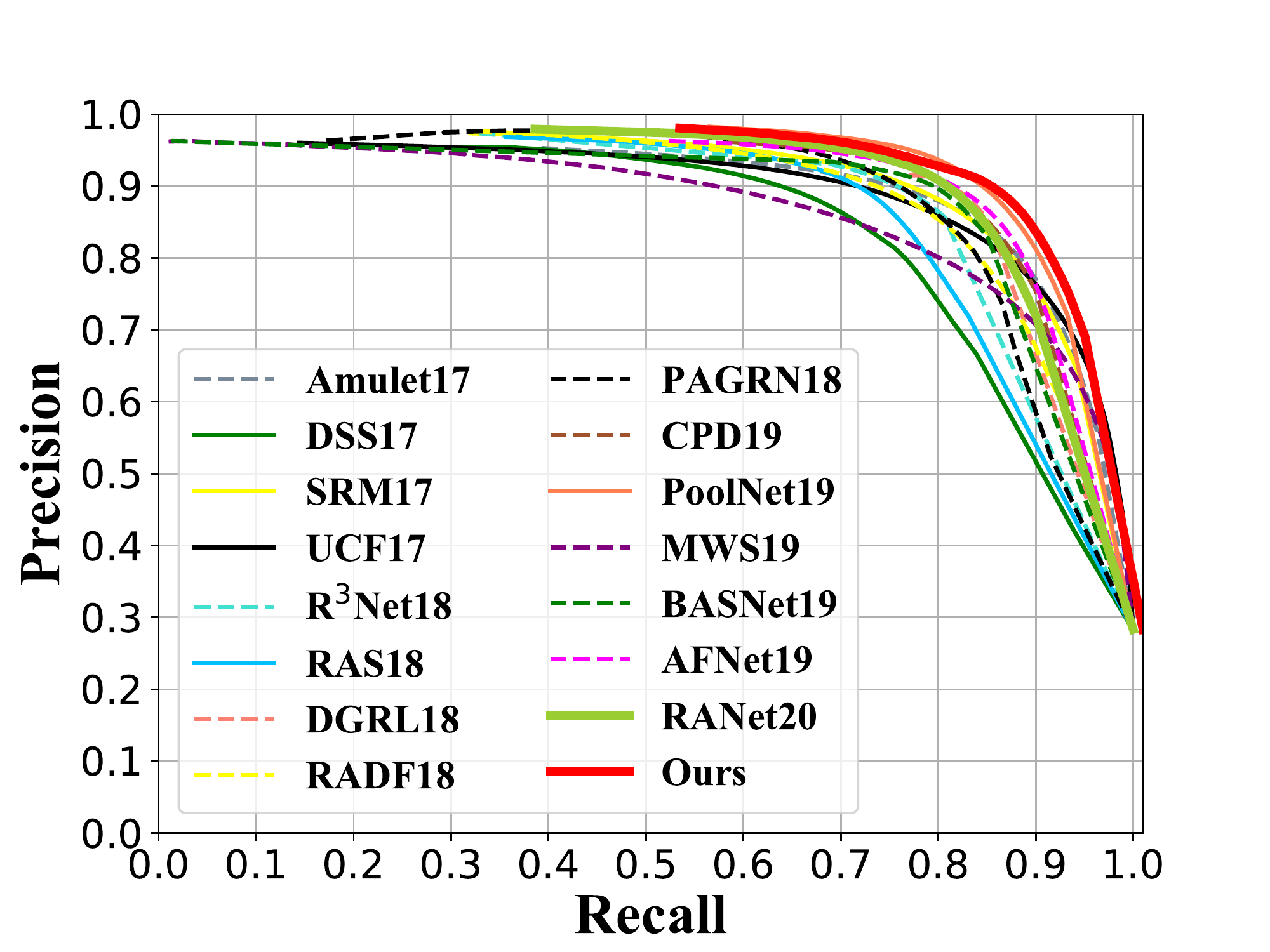}

\end{minipage}}\vspace*{-0.3em}

\subfigure[DUT-OMRON]{
\begin{minipage}[b]{0.205\linewidth}
\includegraphics[width=1\linewidth]{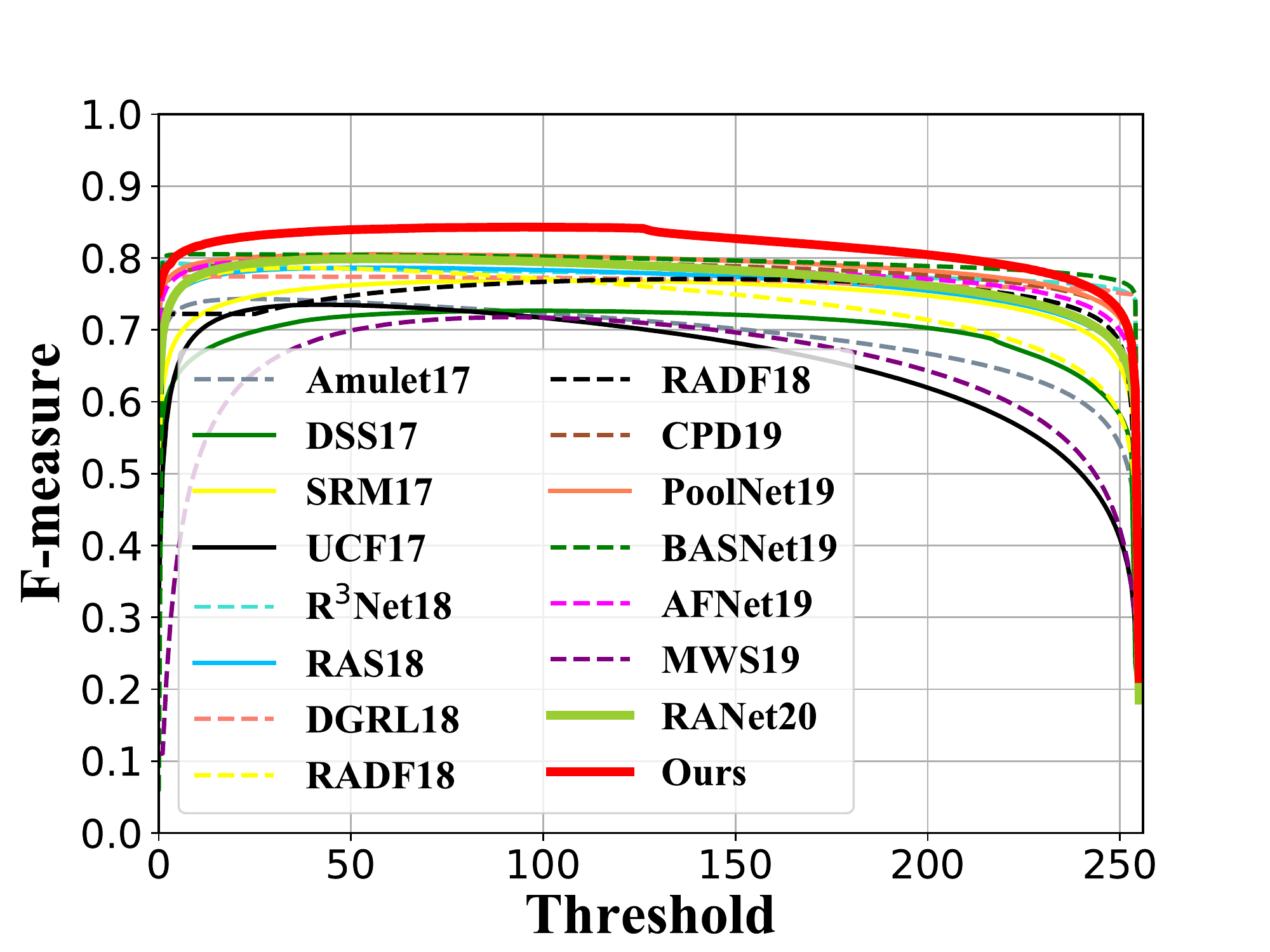}

\end{minipage}}\hspace{0.1pt}\hspace*{-1em}
\subfigure[DUTS-TE]{
\begin{minipage}[b]{0.202\linewidth}
\includegraphics[width=1\linewidth]{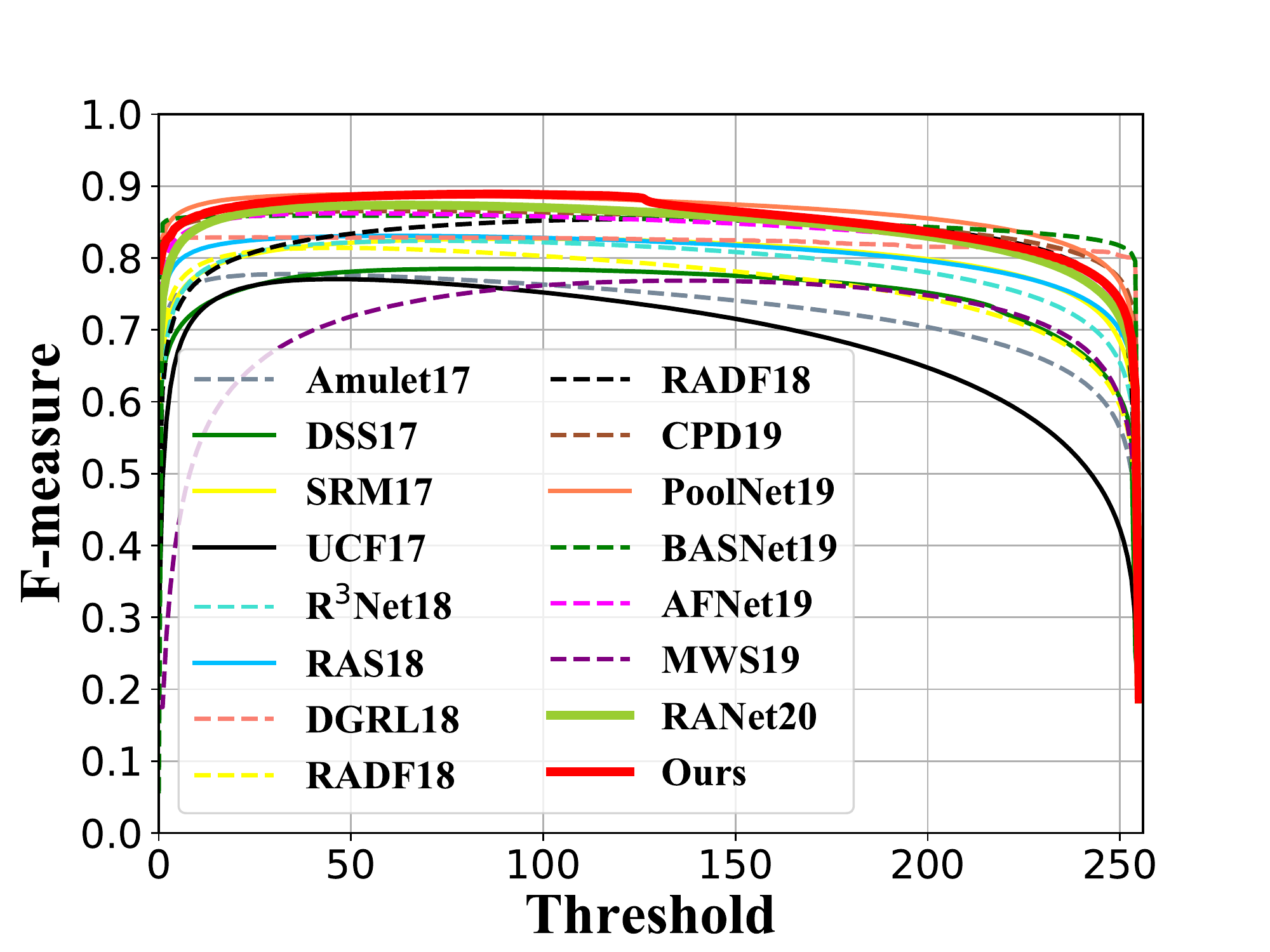}

\end{minipage}}\hspace{0.1pt}\hspace*{-1em}
\subfigure[ECSSD]{
\begin{minipage}[b]{0.202\linewidth}
\includegraphics[width=1\linewidth]{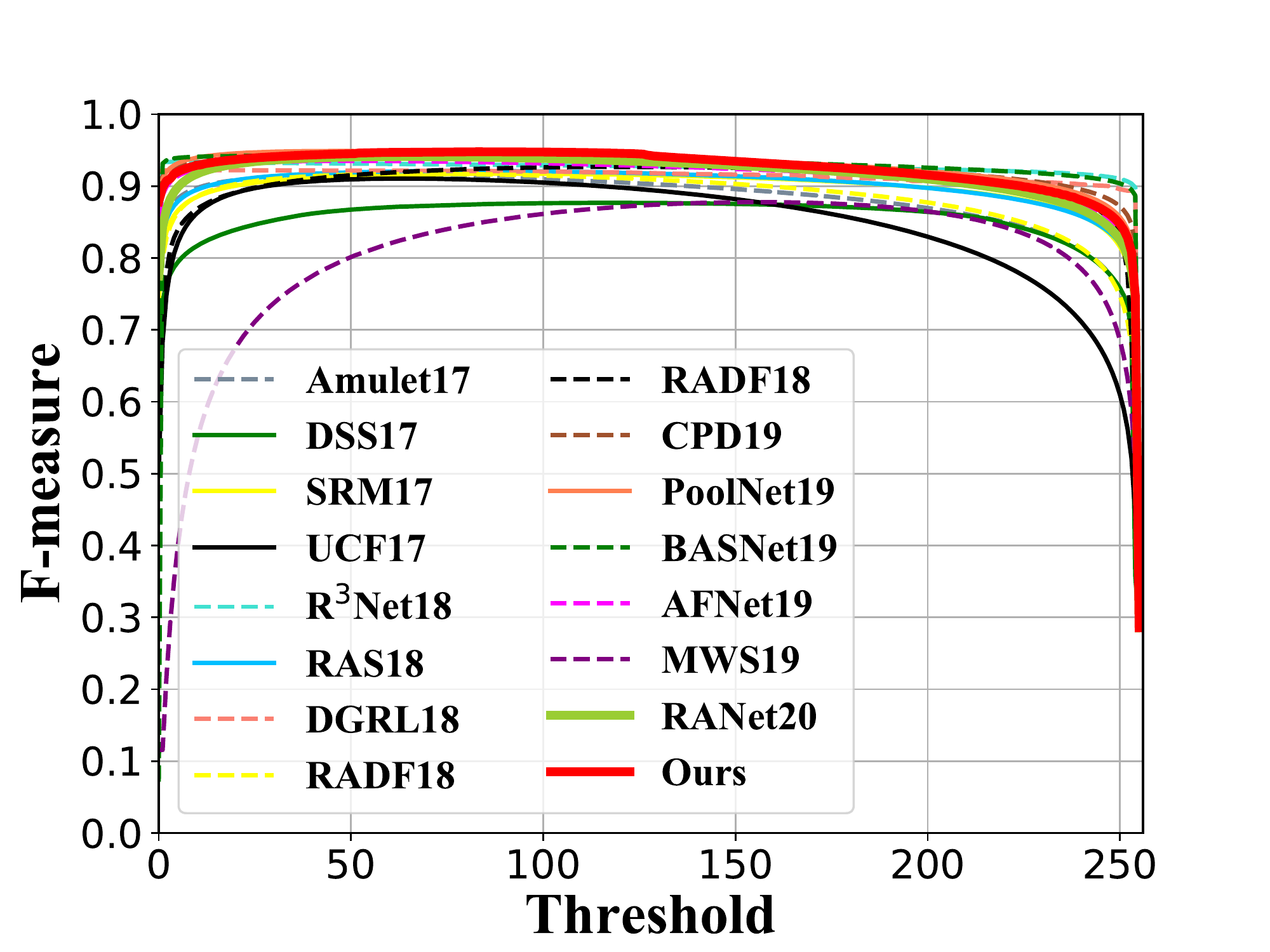}

\end{minipage}}\hspace{0.1pt}\hspace*{-1em}
\subfigure[HKU-IS]{
\begin{minipage}[b]{0.202\linewidth}
\includegraphics[width=1\linewidth]{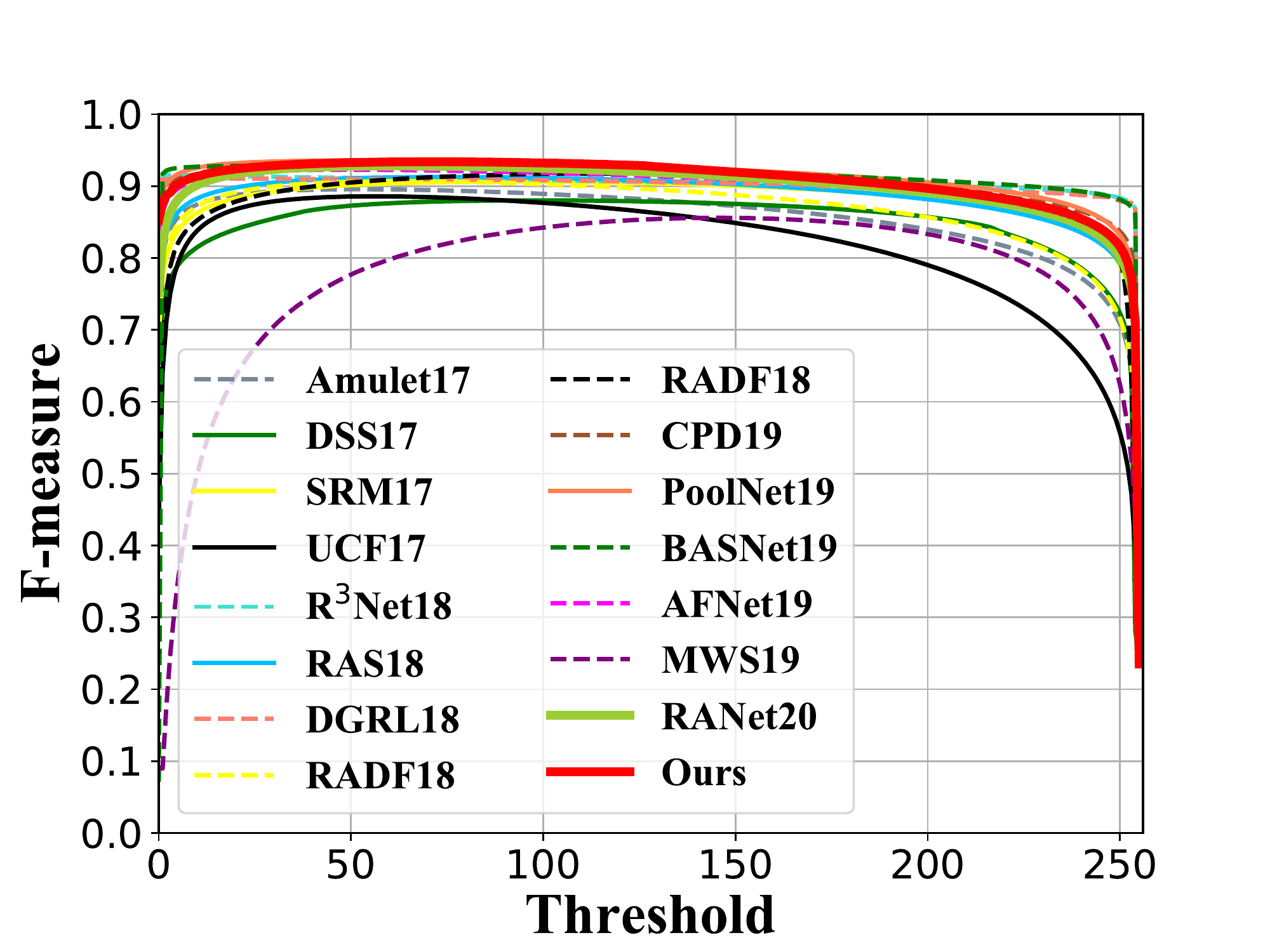}

\end{minipage}}\hspace{0.1pt}\hspace*{-1em}
\subfigure[PASCAL-S]{
\begin{minipage}[b]{0.202\linewidth}
\includegraphics[width=1\linewidth]{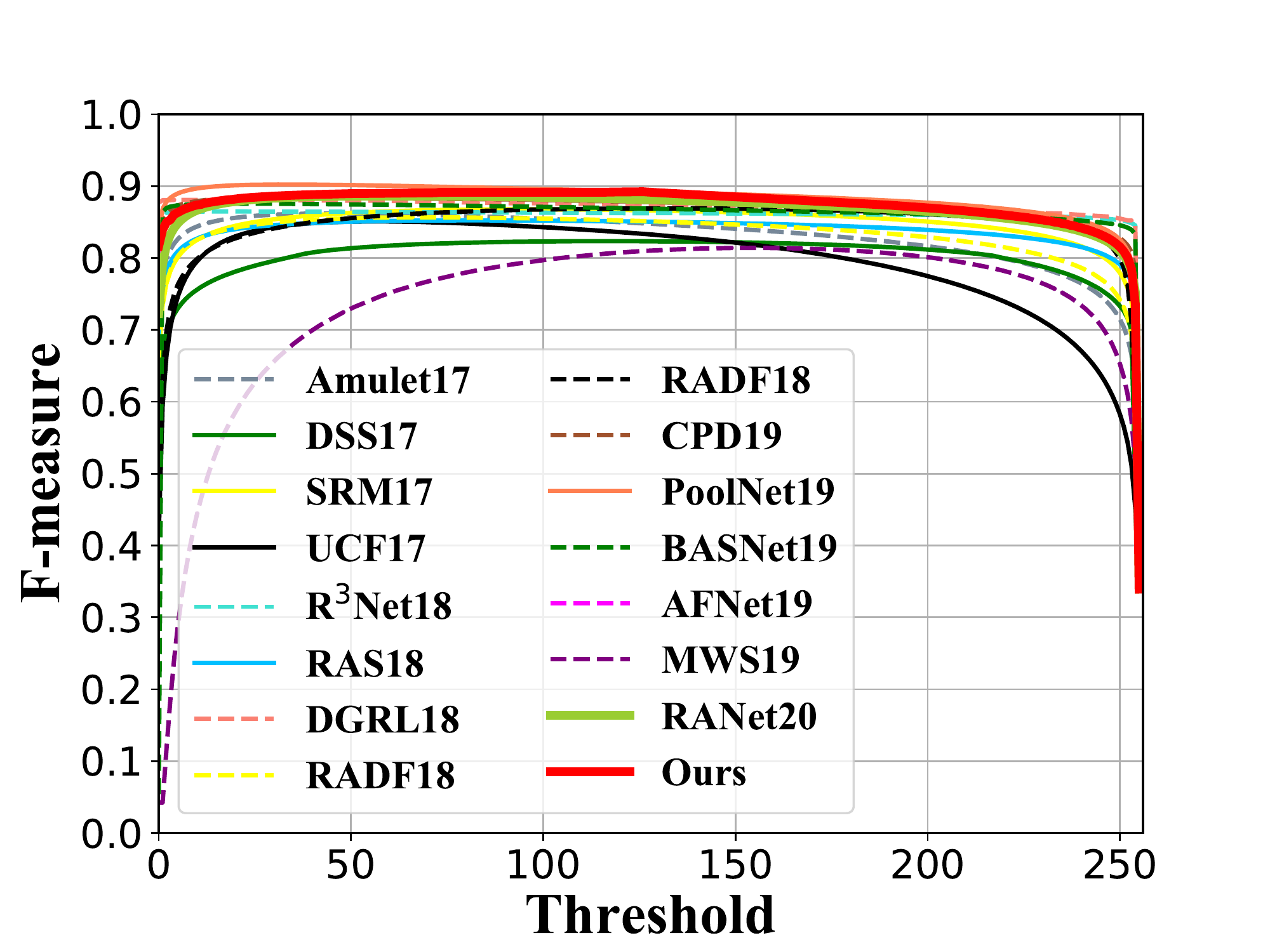}

\end{minipage}}\vspace*{-1em}

\caption{The first row shows the PR curves of the proposed method with other state-of-the-art methods and the second shows F-measure curves. The proposed method performs best among all datasets in terms of all metrics.}
\label{PRcurves11}
\end{figure*}

\vspace{-2.4em}
\subsection{Multi-layer Attention}
\label{multi-layer_attention}
In general, the predicted saliency maps will lose their details if we use sequential scaling operations (e.g., pooling).
Actually, the visual features generated in deep layers are usually abundant in high-level information, while the tiny details are preserved in shallower layers.
Previous works have widely taken full advantage of the multi-level and multi-scale deep features, which introduce features in deep layers to shallower layers via short connections, and this topic has been well studied in~\cite{DSS}.

However, as for our bi-stream network, the overall performance is mainly ensured by the gate mechanism based complementary fusion.
Consequently, the feature map quality in each sub-branch is quite limited, which may result in performance degradation if we follow the conventional ``low$\leftarrow$high'' or ``high$\leftarrow$low'' feature connections directly.

Instead of combining multi-level features indiscriminately, the proposed multi-layer attention (MLA) is developed by using feature maps in deep layers $\textbf{{X}}^R_j$ $(j\in\{4,5\})$, which provide valuable location information for the shallower layers.
We demonstrate the MLA dataflow in Fig.~\ref{method_fig1}, and its details can be formulated as follows:
\begin{equation} \label{method_formulation_9}
  \bm{\alpha}_{j}(l^{'}) = \frac{e^{\bm{\beta}_{j}(l^{'})}}{\sum_{l=1}^{H \times W} e^{\bm{\beta}_{j}(l)}},\ \ \bm{\beta}_j = {tanh(\textbf{X}}^R_j *\textbf{ W} + \bm{b}),
\end{equation}
where $\bm{\beta}_{j} \in  \mathbb{R}^{H \times W}$ integrates the information of all channels in {$\textbf{X}_j^R$}, $\bm{\beta}_{j}(l^{'})$ denotes the feature at location $l^{'}$, and $\bm{\alpha}_{j}$ is the location attention map.
Next, these location attention maps are applied to facilitate those low-level features {$\textbf{{X}}^R_m$}$ (m\in\{1,2\})$ as following:
\begin{equation} \label{method_formulation_10}
   \textbf{X}^R_j \gets f\big(\textbf{X}^R_j,D((\textbf{X}^R_m * \textbf{W} +\textbf{b}) \otimes \bm{\alpha}_j\big)).
\end{equation}
where the function $f(\cdot)$ denotes the element-wise summation, $D(\cdot)$ stands for downsampling operation. After obtaining the updated  $X_j^R$, it will be feed into the decoder part to recover details progressively.
Compared with the widely used multi-scale short-connections, the proposed MLA can improve the overall performance significantly, and the corresponding quantitative proofs can be found in Table~\ref{ablation_studies_tab2}.


\begin{center}
\begin{table*}[!t]
\setlength{\abovecaptionskip}{0pt}%
\setlength{\belowcaptionskip}{5pt}%

\small{
\linespread{2}
\renewcommand\arraystretch{1.1}

\resizebox{1\textwidth}{!}{

\begin{tabular}{|r|c|cc|cc|cc|cc|cc|cc|cc|}
\hline

\multirow{2}{*}{Method}
&\multirow{2}{*}{Backbone}
& \multicolumn{2}{c|}{Training}
& \multicolumn{2}{c|}{DUT-OMRON}
& \multicolumn{2}{c|}{DUTS-TE}
& \multicolumn{2}{c|}{ECSSD}
& \multicolumn{2}{c|}{HKU-IS}
& \multicolumn{2}{c|}{PASCAL-S}
\\
\cline{3-14}
& & Images& Dataset &W-$F_\beta\uparrow$  & S-m$\uparrow$    &W-$F_\beta\uparrow$  & S-m$\uparrow$   &W-$F_\beta\uparrow$& S-m$\uparrow$  &W-$F_\beta\uparrow$  & S-m$\uparrow$  &W-$F_\beta\uparrow$ & S-m$\uparrow$ \\
\hline
\textbf{Ours} &ResNet50+VGG16& 4172& MD4K&{\textcolor{red}{0.761}}&{\textcolor{red}{0.858}} & {\textcolor{red}{0.804}} &  {\textcolor{blue}{0.883}} & {\textcolor{red}{0.915}} &{\textcolor{SeaGreen3}{0.936}} & {\textcolor{red}{0.902}} & {\textcolor{SeaGreen3}{0.921}} & {\textcolor{red}{0.816}}  &  {\textcolor{red}{0.857}} \\


{Ours} &ResNet50+VGG16& 10553& DTS& \textcolor{SeaGreen3}{0.757}& \textcolor{blue}{0.847}& \textcolor{blue}{0.788} & 0.871& \textcolor{SeaGreen3}{0.908} &0.920 & \textcolor{SeaGreen3}{0.893} &\textcolor{blue}{0.914} &\textcolor{SeaGreen3}{0.808} &\textcolor{SeaGreen3}{0.851} \\

{Ours} &ResNet50+VGG16& 10000& MK&0.748 & 0.843 & 0.782 & 0.864 & 0.902 & 0.915 & 0.884 & 0.907 & 0.794 & 0.842\\

{Ours} &ResNet50+ResNet50& 4172& MD4K&{0.723} &{0.834} & {0.782}&  {0.861} & {0.891} &{0.918} & {0.886} & {0.907}  & \textcolor{blue}{0.803} &  {0.848} \\

{Ours} &VGG16+VGG16& 4172& MD4K&{0.716} &{0.831} & {0.780} &  {0.867} & {0.890}&{0.912} & {0.874}& {0.904}  & {0.788} &   {0.102} \\

{RANet20}\cite{RANet20}& VGG16&10553& DTS & 0.671& 0.825&0.743&0.874& 0.866&0.917 & 0.846&0.908 &0.757&0.847\\

{{R$^2$Net20}}\cite{R2Net20} &VGG16&10553 &DTS & -&0.824 & - & 0.861& -& 0.915 & - & 0.903& - &  0.847\\

\textbf{CPD19}$^*$\cite{CPD}   & ResNet50& 4172& MD4K& \textbf{\textcolor{black}0.722}& \textbf{0.845} &\textbf{\textcolor{black}0.785} & \textbf{0.874} & \textbf{\textcolor{black}0.891}& \textbf{0.913} & \textbf{\textcolor{black}0.879}&\textbf{0.912}  &\textbf{\textcolor{black}0.784}& \textbf{0.839}\\

CPD19\cite{CPD}  & ResNet50& 10553& DTS& 0.705 & 0.825 & 0.769 & 0.868 & 0.889& 0.918& 0.866 &0.906 & 0.771 &0.828 \\

\textbf{PoolNet19}$^*$\cite{PoolNet} &   ResNet50&4172& MD4K&\textbf{\textcolor{black} 0.717} &  \textbf{\textcolor{SeaGreen3}{0.851}} &\textbf{{0.786}} &   \textbf{\textcolor{red}{0.894}}& \textbf{{0.893}} & \textbf{\textcolor{red}{0.940}} &\textbf{\textcolor{black}0.885} & \textbf{\textcolor{red}{0.923}} &\textbf{\textcolor{black}0.798} & \textbf{\textcolor{blue}{0.849}}\\

PoolNet19\cite{PoolNet} &ResNet50&10553& DTS& 0.696 & 0.831& 0.775 &\textcolor{SeaGreen3}{0.886} & 0.890 & \textcolor{blue}{0.926}& 0.873 &0.919& 0.781 &0.847 \\

\textbf{AFNet19}$^*$\cite{AFNet} & VGG16&4172 & MD4K&  \textbf{\textcolor{black}0.712} &  \textbf{0.834} &\textbf{\textcolor{black} 0.762} &   \textbf{0.874} &\textbf{\textcolor{black}0.875} &  \textbf{0.916} &\textbf{\textcolor{black}0.863}&  \textbf{0.912} &\textbf{\textcolor{black}0.787} &   \textbf{0.845}\\

AFNet19\cite{AFNet}  &VGG16&10553 & DTS& 0.690 & 0.826& 0.747 & 0.866& 0.867 &0.914& 0.848 & 0.905& 0.772 &0.833  \\

BASNet19\cite{BASNet19} &ResNet34&10553 & DTS& \textcolor{blue}{0.752} & 0.836& \textcolor{SeaGreen3}{0.793} &0.865 & \textcolor{blue}{0.904} & 0.916& \textcolor{blue}{0.889} & 0.909& 0.776 &0.819 \\

MWS19\cite{MWS}  &DenseNet169& 310K&CO+DTS&  0.423& 0.756 & 0.531 &0.757 & 0.652 & 0.828& 0.613 & 0.818& 0.613 &0.753 \\

PAGRN18\cite{PAGRN} &VGG19&10553&DTS& 0.601 & 0.775& 0.685 &0.837 & 0.822 & 0.889& 0.805 & 0.887& 0.701 & 0.793 \\

DGRL18\cite{DGRL} &ResNet50&10553&DTS&  0.709 & 0.806& 0.768 &0.841 & 0.891 & 0.903& 0.875 & 0.895& 0.791 &0.828 \\

RADF18\cite{RADF}   &VGG16&10000&MK& 0.611 & 0.813& 0.635 &0.824 & 0.802 &0.895 & 0.782 &0.888 & 0.709 &0.797  \\

R$^3$Net18\cite{R3Net18} & ResNeXt&10000&MK& 0.726 & 0.817& 0.648 & 0.835& 0.902 &0.910 & 0.877 &0.895 & 0.737 &0.788 \\

SRM17\cite{SRM}  &ResNet50&10553&DTS&0.607 &0.798 & 0.662 &0.835 & 0.825 & 0.895& 0.802 &0.888& 0.736 & 0.817  \\

Amulet17\cite{Amulet} &VGG16&10000&MK& 0.563 &0.781 & 0.594 & 0.803& 0.798 & 0.894& 0.767 &0.883& 0.732 &0.820  \\

UCF17\cite{UCF} &VGG16&10000&MK& 0.465 & 0.758& 0.493 & 0.778& 0.688 & 0.883& 0.656 &0.866 & 0.666 & 0.808 \\

DSS17\cite{DSS} &VGG16& 2500&MB&0.481 & 0.748 & 0.538 &0.790& 0.688 &0.836 & 0.677 &0.852 & 0.626 &0.749   \\

\hline
 \end{tabular}  }}

\caption{The detailed quantitative comparisons between our method and state-of-the-art models in \textbf{weighted F-measure} and \textbf{S-measure}. Top three scores are denoted in \textcolor{red}{red}, \textcolor{SeaGreen3}{green} and \textcolor{blue}{blue}, respectively. \{MD4K, DTS, MK, MB, VOC, TH, CO\} are training datasets which respectively denote \{our small dataset, DUTS-TR, MSRA10K, MSRA-B, PASCAL VOC2007, THUS10K, Microsoft COCO\}. The symbol ``*'' indicates that the target  models were trained on the MD4K dataset.}
\label{results_tab2}
\end{table*}
\end{center}



\section{Experiments and Results}
\label{experiments}
\label{Structure-Meaningful Mid-level Patch Generation}

\subsection{Datasets}
 We have evaluated the performance of the proposed method on five commonly used benchmark datasets, including DUT-OMRON ~\cite{yang2013saliency}, DUTS-TE~\cite{wang2017learning}, ECSSD~\cite{ecssd}, HKU-IS~\cite{zhao2015saliency} and PASCAL-S~\cite{li2014secrets}.
{DUT-OMRON} contains 5,168 high-quality images. Images of this dataset have one or more salient objects with complex backgrounds.
{DUTS-TE} has 5,019 images with high-quality pixel-wise annotations, which is selected from the currently largest SOD benchmark DUTS.
{ECSSD} has 1,000 natural images, which contain many semantically meaningful and complex structures. As an extension of the complex scene saliency dataset, ECSSD is obtained by aggregating the images from  BSD~\cite{martin2004learning} and PASCAL VOC~\cite{everingham2010pascal}.
{HKU-IS}  contains 4,447 images. Most of the images in this dataset have low contrast with more than one salient object.
{PASCAL-S} contains 850 natural images with several objects, which are carefully selected from the PASCAL VOC dataset with 20 object categories and complex scenes.

\subsection{Evaluation Metrics} We have adopted commonly used quantitative metrics to evaluate our method, including the Precision-recall (PR) curves, the F-measure curves, Mean Absolute Error (MAE), weighted F-measure, and S-measure.

{\textbf{PR curves.}} Following the previous settings~\cite{cheng2015global,achanta2009frequency}, we first utilize the standard PR curves to evaluate the performance of our model.

{\textbf{F-measure}}. The F-measure is a harmonic mean of average precision and average recall. we compute the F-measure as
\begin{equation}
\label{distribution difference}
F_{\beta}=\frac{(1 + \beta^2) \times {\rm Precision} \times {\rm Recall}}{\beta^2 \times {\rm Precision + Recall}} ,
\end{equation}
where we set $\beta^2$ to be 0.3 to weigh precision more than recall.

{\textbf{MAE}}. The MAE is calculated as the average pixel-wise absolute difference between the binary ${GT}$ and the saliency map ${S}$ as Eq.~\ref{distribution difference}.
\begin{equation}
\label{distribution difference}
MAE=\frac{1}{ W \times H} \sum \limits_{x=1}^{W} \sum \limits_{y=1}^{H} \Big|{S}(x,y) - {GT}(x,y)\Big| ,
\end{equation}
where $W$ and $H$ are width and height of the saliency map $S$, respectively.

{\textbf{Weighted F-measure}}. Weighted F-measure \cite{w-fmeasure}  define weighted Precision, which is a measure of exactness, and weighted Recall, which is a measure of completeness:

\begin{equation}
\label{distribution difference}
F_{\beta}^w=\frac{(1 + \beta^2) \cdot Precision^w \cdot Recall^w }{ \beta^2 \cdot Precision^w  + Recall^w} ,
\end{equation}

{\textbf{S-measure}}. S-measure \cite{Smeasure}  simultaneously evaluates region-aware $S_r$ and object-aware $S_o$ structural similarity between the saliency map and ground truth. It can be written as follows: $ S_m = \alpha \cdot S_o + (1 - \alpha) \cdot S_r$, where $\alpha$ is set to 0.5.


\begin{figure}[t]
\centering
\includegraphics[width=1\linewidth]{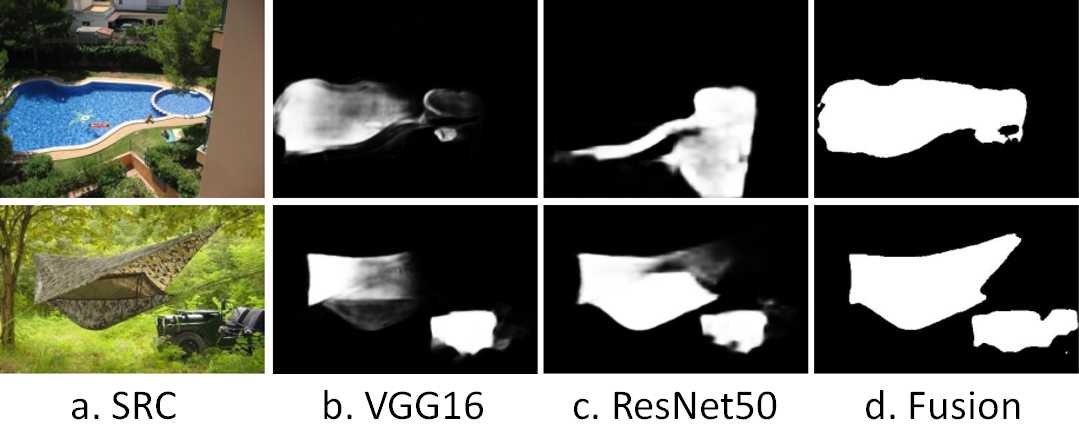}
\caption{Demonstration of the sub-branch complementary status.}
\vspace{-0.2cm}
\label{complementary_results}
\end{figure}



\begin{table}[!t]
\setlength{\abovecaptionskip}{0pt}%
\setlength{\belowcaptionskip}{5pt}%

\renewcommand\arraystretch{1.5}

\Large
\resizebox{1\textwidth}{!}{
\centering
\begin{tabular}{|r|c|c|c|c|c|}

\hline
Method& Model(MB)&Encoder(MB)&Decoder(MB)& FLOPs(G) & Params(M)   \\
\hline
Ours &235.5&152.6&82.9&65.53 &71.67 \\
\hline
{CPD19\cite{CPD}} & 192 & 95.6& 96.4 &17.75& 47.85\\
\hline
BASNet19\cite{BASNet19} & 348.5 & 87.3 &261.2 &127.32& 87.06 \\

\hline
{PoolNet19\cite{PoolNet}}  & 278.5 & 94.7&183.8 &88.91& 68.26 \\
\hline

\end{tabular}}

\caption{The number of model size, FLOPs and parameters comparisons of our method with 3 state-of-the-art models.}
\label{model_size_analysis}
\end{table}

\vspace{-0.4cm}
\subsection{Implementation Details}
The proposed method is developed on the public deep learning framework PyTorch. We run our model in a machine with an i7-6700 CPU (3.4 GHz and 8 GB RAM) and a NVIDIA GeForce GTX 1070 GPU (with 8G memory). Our bi-stream model was trained on the proposed small training dataset (MD4K). Then, we test our model on the other datasets. Due to the GPU memory limitation, we set the mini-batch size to 4. We use the stochastic gradient descent (SOD) method to train our model with a momentum 0.99 and weight decay 0.0005. We use the fixed learning rate policy and set the base learning rate to $10^{-10}$. Learning stops after 30K iterations, and we use standard Binary Cross Entropy loss during learning.

\subsection{Comparison with the state-of-the-art Methods}
We have compared our method with 16 state-of-the-art models, including
 DSS17~\cite{DSS}, Amulet17~\cite{Amulet}, UCF17~\cite{UCF}, SRM17~\cite{SRM}, R$^3$Net18\cite{R3Net18}, RADF18~\cite{RADF}, PAGRN18~\cite{PAGRN}, DGRL18~\cite{DGRL}, MWS19 \cite{MWS}, CPD19 \cite{CPD}, AFNet19 \cite{AFNet}, PoolNet19 \cite{PoolNet}, BASNet19\cite{BASNet19}, {{R$^2$Net20}}\cite{R2Net20}, {MRNet20}\cite{MRNet20} and RANet20\cite{RANet20}. For all of these methods, we use the original codes with recommended parameter settings or the saliency maps provided by the authors. Moreover, our results are diametrically generate by model without relying on any post-processing and all the predicted saliency maps are evaluated with the same evaluation code.

\begin{figure}[t]
\centering
\includegraphics[width=3.3in]{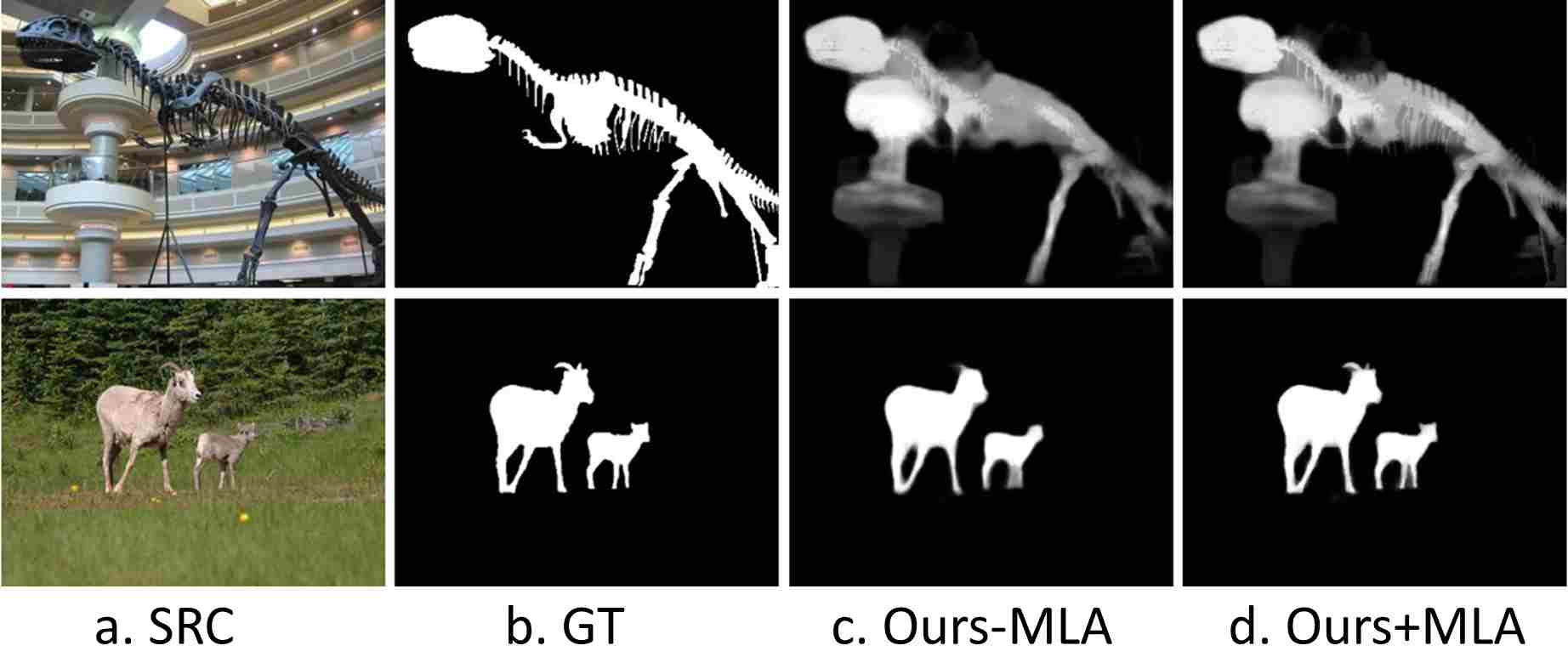}
\caption{Visual comparison of the proposed model with multi-layer attention (``Ours+MLA'') and without multi-layer attention (``Ours-MLA'').}
\vspace{-0.2cm}
\label{ablation_fig1}
\end{figure}


\begin{table}[!t]
\setlength{\abovecaptionskip}{0pt}%
\setlength{\belowcaptionskip}{5pt}%

\renewcommand\arraystretch{1.3}

\resizebox{1\textwidth}{!}{
\centering
\begin{tabular}{|c|c|c|c|c|c|}

\hline
Method& Ours &RANet20 &R$^2$Net20  & MRNet20 & BASNet19 \\

\hline
FPS & \textbf{\textcolor{red}{23}} &42& 33 &14& 25 \\

\hline
Method& CPD19 & PoolNet19  &AFNet19 & DGRL18&  RADF18 \\

\hline
FPS & 62 & 27& 23 & 6 & 18 \\
\hline
\end{tabular}
}
\caption{Running time comparisons.}
\label{time_analysis_tab1}
\end{table}

\vspace{0.2cm}
\noindent \textbf{Quantitative Comparisons}.
As a commonly used quantitative evaluation venue, we first investigate our model using the PR curves. As shown in the first row of Fig. \ref{PRcurves11}, our model can consistently outperform the state-of-the-art models on all tested benchmark datasets. Specifically, the proposed model outperforms other models on DUT-OMRON datasets.
Meanwhile, our model also is evaluated by F-measure curves as shown in the second row of Fig. \ref{PRcurves11}, which also demonstrates the superiority of our method.
The detailed F-measure, MAE, weighted F-measure and S-measure  values are provided in Table~\ref{results_tab1} and Table \ref{results_tab2}, in which our method also performs favorably against other state-of-the-art approaches.

\vspace{0.2cm}
\noindent \textbf{Qualitative Comparisons}.
We demonstrate the qualitative comparisons in Fig.~\ref{result_maps1}.
The proposed method not only detects the salient objects accurately and completely, but preserves subtle details also.
Specifically, the proposed model can adapt to various scenarios as well, including the object occlusion case (raw 1), the complex background case (row 2), the small object case (row 3) and the low contrast case (row 4).
Moreover, our method can consistently highlight the foreground regions with sharp object boundaries.

To further illustrate the complementary status between VGG16 and ResNet50, Fig.~{\ref{complementary_results}} shows the saliency maps of these two sub-branches in mining salient regions. We observe that these two sub-branches are capable of revealing different but complementary salient regions.


\begin{center}
\begin{table}[!t]
\setlength{\abovecaptionskip}{0pt}%
\setlength{\belowcaptionskip}{-0.5em}%

\scriptsize
\resizebox{1\textwidth}{!}{

\renewcommand\arraystretch{1.1}
\begin{tabular}{|r|cc|cc|cc|cc|}

\hline
\multirow{2}{*}{Fusion Method}
& \multicolumn{2}{c|}{DUT-OMRON}
& \multicolumn{2}{c|}{DUTS-TE}
& \multicolumn{2}{c|}{ECSSD}

\\
\cline{2-7}
&  max$F_\beta$ & MAE  &max$F_\beta$ & MAE  &  max$F_\beta$ & MAE   \\
\hline
\rowcolor{mygray}
Conv w/ GCN (Ours)    &\textbf{\textcolor{red}{0.857}}&\textbf{\textcolor{red}{0.044}}  &\textbf{\textcolor{red}{0.884}}&  \textbf{\textcolor{red}{0.038}} & \textbf{\textcolor{red}{0.945}} &\textbf{\textcolor{red}{0.036}}  \\

Conv w/ GCN (LSTM) &{0.834} &{0.046} & {0.864}&  {0.045} & {0.934} &{0.042}  \\

Conv w/o GCN   &  0.821& 0.049 &  {0.844} &{0.051} &  0.927& {0.048}  \\
\cline{1-7}
\rowcolor{mygray}
Sum w/ GCN   &  \textbf{\textcolor{black}{ {0.848}}} &   \textbf{\textcolor{black}{{0.047}}} &  \textbf{\textcolor{black}{ 0.873}} &  \textbf{\textcolor{black}{0.044}} &  \textbf{\textcolor{black}{{0.925}}}&   \textbf{\textcolor{black}{0.043}} \\

Sum w/o GCN   &  0.813& 0.055 &  {0.845} &{0.052} &  0.897& {0.049}  \\

\cline{1-7}
\rowcolor{mygray}
Concat w/ GCN    &  \textbf{\textcolor{black}{ 0.827}}&  \textbf{\textcolor{black}{0.049 }}&   \textbf{\textcolor{black}{{0.862}}} & \textbf{\textcolor{black}{{0.047}}} &  \textbf{\textcolor{black}{ 0.908}}&  \textbf{\textcolor{black}{{0.046}}}  \\

Concat w/o GCN     &  0.802& 0.059 &  {0.847} &{0.058} &  0.887& {0.054}   \\

\cline{1-7}
\rowcolor{mygray}
Max w/ GCN   &  \textbf{\textcolor{black}{0.818}}& \textbf{\textcolor{black}{0.050}} &  \textbf{\textcolor{black}{{0.853}}} &\textbf{\textcolor{black}{{0.048}}} & \textbf{\textcolor{black}{ 0.909}}& \textbf{\textcolor{black}{{0.047}}}   \\

Max w/o GCN &  {0.813}& 0.054 &  {0.836} &{0.054} &  0.887& {0.053}  \\

\hline

 \end{tabular}  }

\caption{Performance comparisons of different fusion strategies, where ``w/'' denotes ``with'', ``w/o'' denotes ``without''; GCN: Gate Control Unit; Conv, Sum, Concat, Max  are four conventional fusion schemes mentioned in Sec.~\ref{gate_control_unit}. ``Conv w/ GCN (LSTM)'' denotes the performance using the gate control logic of LSTM.}
\label{ablation_studies_tab1}
\end{table}
\end{center}

\vspace{-0.4cm}

\noindent \textbf{Running Time and Model Complexity Comparisons}.
Table~{\ref{time_analysis_tab1}} shows the running time comparisons. This evaluation was conducted on the same machine with an i7-6700 CPU and a GTX 1070 GPU, in which our model achieves 23 FPS. Furthermore, we compare model size, FLOPs and the number of parameters with other popular methods in Table~\ref{model_size_analysis}. In spite of using two feature extractors, our model complexity is not so much heavy and only slightly worse than CPD \cite{CPD}. As shown in Table~\ref{model_size_analysis}, previous methods  treat the feature backbones as the off-the-shelf tools and pay more attention to design complex decoder to improve the overall performance . In sharp contrast, the propose bi-stream network is concentrate on the encoder instead of devising a complex decoder and achieves new state-of-the-art performance, showing the importance of feature extractor.

\subsection{Component Evaluations}

\noindent  \textbf{Effectiveness of the Proposed MD4K Dataset}. To illustrate the advantages of the proposed dataset, we train the proposed bi-stream network on MD4K and DUTS-TR datasets respectively. Compared to train on the DUTS-TR dataset, the bi-stream network with the MD4K dataset achieves better performance in terms of different measures, which demonstrates the effectiveness of the proposed dataset. Besides,
as shown in the rows 9-14 of Table~\ref{results_tab1}, three state-of-the-art methods (i.e., PoolNet19, CPD19 and AFNet19) are trained on either the DUT-OMRON dataset or our MD4K dataset respectively.
Clearly, models trained on the MD4K dataset achieve better performance than the ones trained on the large-scale DUT-OMRON dataset, also showing the effectiveness of the proposed MD4K dataset.
To demonstrate the importance of balanced semantic distribution, except for the proposed bi-stream network, we also train 3 state-of-the-art models on M4K and D4K which is randomly selected from MSRA10K and DUTS-TR respectively as shown in Table~\ref{ablation_studies_tab2}. There is no exception, models trained on semantic balanced datasets achieves significantly improve their performance. The primary reason is that models, trained on a semantical category balanced dataset, make itself learned on more practical scenes and consequently will enhance generability of model to other datasets.

\begin{center}
\begin{table}[!t]
\setlength{\abovecaptionskip}{0pt}%
\setlength{\belowcaptionskip}{-1em}%

\scriptsize
\resizebox{1\textwidth}{!}{

\renewcommand\arraystretch{1.1}
\begin{tabular}{|r|cc|cc|cc|cc|}

\hline
\multirow{2}{*}{Method}
& \multicolumn{2}{c|}{DUT-OMRON}
& \multicolumn{2}{c|}{DUTS-TE}
& \multicolumn{2}{c|}{ECSSD}

\\
\cline{2-7}
&  max$F_\beta$ & MAE  &max$F_\beta$ & MAE  &  max$F_\beta$ & MAE   \\
\hline
\rowcolor{mygray}
Ours(MD4K)    &\textbf{\textcolor{red}{0.857}}&\textbf{\textcolor{red}{0.044}}  &\textbf{\textcolor{red}{0.884}}&  \textbf{\textcolor{red}{0.038}} & \textbf{\textcolor{red}{0.945}} &\textbf{\textcolor{red}{0.036}}   \\

Ours(D4K) & 0.825 & 0.048 & 0.838 &  0.051 & 0.905 &  0.048 \\

Ours(M4K) & 0.820 & 0.060 & 0.823 &  0.052 & 0.887 &  0.050   \\
\hline
\rowcolor{mygray}
CPD19(MD4K) & \textbf{ 0.762} & \textbf{0.052} & \textbf{0.850} & \textbf{0.040} & \textbf{0.943} & \textbf{0.037}  \\

CPD19(D4K) &  0.721 &  {0.063}& 0.824 & 0.048 & 0.902 & 0.043  \\

CPD19(M4K) &  0.722 &  0.060& 0.818 & 0.056 & 0.889 & 0.061 \\

\hline
\rowcolor{mygray}
PoolNet19(MD4K) &  \textbf{0.767} &  \textbf{0.051} & \textbf{0.863} &   \textbf{0.042} & \textbf{0.931} & \textbf{0.040} \\

PoolNe19(D4K) & 0.738 & 0.064& 0.839 &  0.047 & 0.907 & {0.043}  \\

PoolNet19(M4K) & 0.733  & 0.065 & 0.836 &  0.048 & {0.897} & 0.045   \\

\hline
\rowcolor{mygray}
AFNet19(MD4K) &   \textbf{0.765} &  \textbf{0.054} &  \textbf{0.842} &   \textbf{0.044} &  \textbf{0.932} &  \textbf{0.041}  \\
AFNet19(D4K) &  0.737 &  0.065 & 0.823 & 0.057 & 0.891 & 0.062 \\

AFNet19(M4K) & 0.728 & 0.063 &  0.830& 0.053 & 0.895  & 0.060  \\

\hline
\rowcolor{mygray}
 w/ MLA     &\textbf{\textcolor{red}{0.857}}&\textbf{\textcolor{red}{0.044}}  &\textbf{\textcolor{red}{0.884}}&  \textbf{\textcolor{red}{0.038}} & \textbf{\textcolor{red}{0.945}} &\textbf{\textcolor{red}{0.036}}   \\

 w/o MLA  &   0.834 & 0.050& 0.858 &  0.043 &  0.923 &  0.044  \\

\hline
 \end{tabular}  }
\caption{Quantitative proofs regarding the effectiveness of our proposed small-scale training set (MD4K), where D4K (M4K) represents randomly extract 4172 images from DUTS-TR (MSRA10K) datasets.}
\label{ablation_studies_tab2}
\end{table}
\end{center}

\vspace{-0.4cm}
\noindent  \textbf{Effectiveness of the Proposed Bi-stream Network}. To demonstrate the effectiveness of the proposed bi-stream network,
we also implement the proposed bi-stream network by using other sub-network combinations, i.e., ``VGG16+VGG16'' and ``ResNet50+ResNet50'', see Table~\ref{results_tab1}.
Compared to the ``VGG16+VGG16'' and ``ResNet50+ResNet50'' based model, which trained on the MD4K dataset, the proposed bi-stream network achieves better performance. In addition, we also report the performance of the proposed bi-stream network trained on the DUTS-TR dataset as shown in 2nd row of Table \ref{results_tab1}. As we can see, our model trained on DUTS-TR achieves better performance than state-of-the-art models, which also suggests that the proposed bi-stream network is effective.

\noindent  \textbf{Effectiveness of the Gate Control Unit}.
To validate the exact contribution of the proposed Gate Control Unit (GCN), we first tested previously mentioned 4 fusion schemes (Sec.~\ref{gate_control_unit}) without using our GCN as the baselines.
Then, we apply our GCN into these conventional fusion schemes, and the corresponding quantitative results can be found in Table~\ref{ablation_studies_tab1}, in which our GCN can boost the conventional fusion schemes significantly.

\noindent  \textbf{Effectiveness of the Multi-layer Attention}.
As shown in the last row of Table~\ref{ablation_studies_tab2}, the overall performance constantly improves after using the multi-layer attention, e.g., F-measure: $0.834\rightarrow0.857$, MAE: $0.05\rightarrow0.044$ on the DUT-OMRON dataset. Additionally, Fig. \ref{ablation_fig1} shows that the proposed multi-layer attention is capable of sharping the object boundaries.



\section{Conclusion}
\label{conclusion}
In this paper, we have provided a deeper insight into the interrelationship between the SOD performance and the training dataset, including the choice of training dataset and the amount of training data that the model requires. Inspired by our findings, we have built a small, hybrid, and scene category balanced training dataset to alleviate the demands for the large-scale training set. Moreover, the proposed training set can essentially improve the state-of-the-art methods performances, providing a paradigm regarding how to effectively design a training set.
Meanwhile, we have proposed a novel bi-stream architecture with gate control unit and multi-layer attention to take full advantage of the proposed small-scale training set.
Extensive experiments have demonstrated that the proposed bi-stream network can work well with the small training set, achieving new state-of-the-art performance on five benchmark datasets.

\setcounter{equation}{0}   
\renewcommand{\theequation}{A\arabic{equation}} 
\setcounter{figure}{0}   
\renewcommand{\thefigure}{A\arabic{figure}} 

\ifCLASSOPTIONcaptionsoff
  \newpage
\fi
\bibliographystyle{IEEEtran}
\bibliography{reference}

\end{document}